\pdfoutput=1

\documentclass[11pt]{article}

\usepackage{ACL2023}
\usepackage{float}
\usepackage{multirow}
\usepackage{listingsutf8}
\usepackage{stackengine}
\usepackage{subcaption}
\usepackage{longtable}
\usepackage{graphicx}
\usepackage{changepage} 
\usepackage{latexsym}
\usepackage[utf8]{inputenc} 
\usepackage[T1]{fontenc}

\usepackage[T1]{fontenc}
\usepackage{times}

\usepackage{microtype}

\usepackage{inconsolata}

\title{From LLM to NMT: Advancing Low-Resource Machine Translation with Claude}

\author{Maxim Enis \and Mark Hopkins\\
  Williams College \\
  \texttt{me4@williams.edu} \\}

\begin{document}
\maketitle
\begin{abstract}
We show that Claude 3 Opus, a large language model (LLM) released by Anthropic in March 2024, exhibits stronger machine translation competence than other LLMs. Though we find evidence of data contamination with Claude on FLORES-200, we curate new benchmarks that corroborate the effectiveness of Claude for low-resource machine translation into English. We find that Claude has remarkable \textit{resource efficiency} -- the degree to which the quality of the translation model depends on a language pair's resource level.  Finally, we show that advancements in LLM translation can be compressed into traditional neural machine translation (NMT) models. Using Claude to generate synthetic data, we demonstrate that knowledge distillation advances the state-of-the-art in Yoruba-English translation, meeting or surpassing strong baselines like NLLB-54B and Google Translate.
\end{abstract}

\section{Introduction}
\label{sec:intro}

Large language models (LLMs) \cite{kaplan2020scaling, hoffmann2022training} have emerged as a breakthrough technology for natural language processing. LLMs have demonstrated a remarkable ability to perform many downstream tasks \cite{brown2020language}, including machine translation. In fact, \citet{zhu2023multilingual, jiao2023chatgpt, robinson-etal-2023-chatgpt} have shown that the translation performance of GPT-4 \cite{achiam2023gpt} is competitive with state-of-the-art neural machine translation (NMT) systems on several high-resource language pairs. However on low-resource language pairs, they showed that existing LLMs lag behind specialized systems like Meta AI's NLLB-54B\footnote{Model card: \href{https://huggingface.co/facebook/nllb-moe-54b}{https://huggingface.co/facebook/nllb-moe-54b}.} \cite{nllbteam2022language}.

In this paper, we present evidence that this performance gap may be closing. On 25\% of evaluated language pairs, we show that Claude 3 Opus, an LLM produced by Anthropic, surpasses strong baselines, like Google Translate and NLLB-54B, when translating into English. Surprisingly, the source languages range from very low- to high-resource, indicating that Claude may have broader machine translation capabilities than prior LLMs. Our findings are based on newly created parallel corpora which are verifiably unseen by Claude, since an auxiliary finding of our paper is that Claude exhibits evidence of data contamination \citep{sainz-etal-2023-nlp} on existing benchmarks like FLORES-200 \citep{ flores-two, flores101, nllbteam2022language}.

We also corroborate the findings of \citet{zhu2023multilingual}, who have shown that current LLMs are most effective at machine translation when English is the target language (i.e. they are better at \verb|xxx->eng| translation than \verb|eng->xxx| translation). Although Claude outperforms NLLB-54B on 55.6\% of language pairs in the \verb|xxx->eng| direction, it only outperforms NLLB-54B on 33.3\% of language pairs in the \verb|eng->xxx| direction, suggesting that supervised baselines still have an edge over LLMs when English is \textbf{not} the target language of the translation task.

However, the costs and inference time of massive LLMs like Claude limit the scope of their applicability for machine translation. In section \ref{sec:knowledge-distillation}, we show that distillation techniques \cite{hinton2015distilling, kim2016sequencelevel} can be applied productively to LLMs to create compact NMT models that outperform the state-of-the-art. We believe that further refinements and optimizations of our methods can result in even better performance, and that many more language pairs, whether currently supported by translation systems or not, are amenable to our approach.

In summary, we make the following contributions:
\begin{enumerate}
    \item We find evidence of data contamination with Claude on the FLORES-200 benchmark.
    \item By creating new and unseen evaluation benchmarks for 36 language pairs, we show that Claude nonetheless demonstrates state-of-the-art machine translation ability for many language pairs, including low- and very low-resource language pairs. We provide evidence that Claude's translation performance (when English is the target language) has higher \textit{resource efficiency} than other LLMs.
    \item We show that when translating from English into a low-resource language, a large gap still exists between LLMs and state-of-the-art neural machine translation (NMT) systems on most languages. Even so, we show that Claude outperforms strong baselines for \textbf{two} such language pairs.
    \item We demonstrate that translation abilities of Claude can be leveraged to advance the state-of-the-art in traditional neural machine translation (NMT) by generating a parallel corpus from Claude translations and fine-tuning the inexpensive model on this corpus. We describe an approach that leverages Claude's context window to reduce distillation costs and improve translation quality, by `batching'' sentences from the same web-crawled document into the same prompt.
\end{enumerate}

\section{Background}
\label{sec:background}

\paragraph{LLM translation.} Prior work has examined the translation abilities of large language models. \citet{robinson-etal-2023-chatgpt} and \citet{zhu2023multilingual} both run empirical studies assessing the translation ability of LLMs like GPT-4 on the FLORES-200 benchmark. Both works find that some LLMs are competitive with NLLB-54B on high-resource languages, but lag behind on low-resource languages. \citet{robinson-etal-2023-chatgpt} find that the \textit{number of Wikipedia pages in a given language} is the most important feature to predict the performance of GPT-4 translation. \citet{stap-araabi-2023-chatgpt} evaluate GPT-4 translation of low-resource indigenous American languages into Spanish. For all languages considered, the LLM underperforms a fine-tuned, supervised multilingual NMT model. We will reexamine these results in light of the release of Claude 3 Opus.

\paragraph{Dataset contamination.} \citet{zhu2023multilingual} examine dataset contamination for LLMs, finding that FLORES-200 is unsuitable for evaluation on the BLOOMZ \citep{muennighoff-etal-2023-crosslingual} LLM. \citet{sainz-etal-2023-nlp} highlight the dangers of evaluating closed-source LLMs on public benchmarks, arguing that data leakage should be a central concern for modern natural language processing researchers. \citet{carlini2023quantifying} also consider the problem of quantifying data contamination on closed-source LLMs, using an \textit{information extraction} approach to detect contamination.

\paragraph{Knowledge distillation with LLMs.} To the best of our knowledge, \citet{li2024mtpatcher} is the only work to examine knowledge distillation between LLMs and NMT systems. Their approach involves identifying corrections to translations generated by a student model, making those corrections, and then synthetically generating similar parallel sentences. Although their approach shows promise in the high-resource Chinese-English and English-German language pairs, we note that synthetic corpus generation from monolingual data is inefficient on high-resource pairs due to the large quantity of data needed to achieve an increase in model performance. In our paper, we examine knowledge distillation into low-resource language pairs, which is a substantially less data-hungry setting.

Relatedly, \citet{DBLP:journals/corr/abs-1910-01348} studied knowledge distillation from very large models to much smaller models. By running knowledge distillation experiments on CNNs evaluated on the ImageNet dataset \citep{5206848}, the authors conclude that ``bigger models are not better teachers'', and that higher teacher accuracy does not necessarily correspond to better child model performance. Given these negative results, we aim to find whether distillation is viable from LLMs to much smaller NMT models.

\paragraph{LLM-based document translation.} We introduce a novel approach to generate training data by using sentence-aligned document translation. \citet{karpinska-iyyer-2023-large} have evaluated various approaches to document translation with LLMs, such as paragraph-level translation and sentence-level translation with the paragraph context. Neither of these approaches support our need for single-prompt sentence-by-sentence document translation. \citet{ wang-etal-2023-document-level} prompt ChatGPT to translate a document sentence-by-sentence by including sequential boundary tags, but they find that ChatGPT tends translate ``without adhering to strict sequential boundaries'', making it difficult to extract parallel sentence pairs. 

\section{Experiments}
\label{sec:experiments}

Our main experiments involve testing a variety of different languages, from high- to low- to very low-resource, against a number of different datasets.

\subsection{Languages}
Following \cite{koishekenov-etal-2023-memory}, we classify languages as very low-resource if they have less than 100k bitexts, low-resource if they have between 100k and 1m bitexts, and high-resource if they have more than 1m bitexts, according to \cite{nllbteam2022language}\footnote{See \href{https://tinyurl.com/535f7ust.}{https://tinyurl.com/535f7ust}.}.
We experiment on English and a selection of 36 other languages, of which 15 are high-resource, 17 are low-resource, and 4 are very low-resource. All languages are supported by Google Translate, NLLB-200 \citep{flores-two}, and are included in the FLORES-200 dataset. Every language is evaluated in both the \verb|eng->xxx| and \verb|xxx->eng| directions. We do not conduct experiments on non-English-centric language pairs. The full list of languages is provided in Table \ref{tab:language-table}.

\subsection{Datasets}
We benchmark our model against the following datasets.

\subsubsection{FLORES-200}
FLORES-200 is a high-quality evaluation dataset containing human-curated translations between English and 204 different languages \citep{ flores-two, flores101, nllbteam2022language}. The dataset serves as a universal benchmark for which to evaluate state-of-the-art in a number of different languages. Due to budget constraints\footnote{The cost comes mainly from requests to the Claude API.}, we preselect 100 random sentences from the FLORES devtest split and test each language pair on this subset. Empirically, we find that NLLB-54B performance on this subset is consistent with the published metrics\footnote{See \href{https://tinyurl.com/nllb200moe54bmetrics}{https://tinyurl.com/nllb200moe54bmetrics}.}.

\subsubsection{BBC News}
The FLORES datasets are high-quality, but might have both source and target sentences seen by the LLM. Thus data contamination is possible. We note that even the private FLORES-200 test set\footnote{Only the \texttt|dev| and \texttt|devtest| splits have been publicly released.} may be subject to this bias for LLM evaluation, since the English data comes from Wikipedia, on which the LLM has almost certainly been trained. These challenges highlight the importance of developing a machine translation benchmark for LLMs with unseen source and target sentences.

In order to check that this bias does not influence the results, we automatically create totally unseen datasets by aligning articles on BBC News. The concept of bitext mining is not novel \citep{heffernan-etal-2022-bitext}; however, typically the quality of mined data is insufficient for evaluation. Using heuristics specific to BBC, we improve the confidence in aligned translations. For example, BBC articles that are translations of each other tend to include the same images, so we restrict parallel mining to documents with similar images according to Google Reverse Image Search\footnote{\href{images.google.com}{images.google.com}}. A more detailed explanation of the mining approach is provided in Appendix \ref{sec:appendix:bbc-mined}.

Through our parallel mining approach, we find parallel sentences from news articles between English and 36 other languages. Then, we filter all articles with publication date prior to the training cutoff of Claude, and finally evaluate on a subset of 100 sentences from each language.

\subsubsection{Maltese Speech}
In order to verify the LLM performance on different domains, we consider another totally unseen dataset: Maltese-English speech pairs taken from transcriptions and translations of the Common Voice, created for the shared IWSLT 2024 task \citep{hernandez-mena-etal-2020-masri}. We test on a random selection of 100 sentences from this dataset.

\subsubsection{Evaluation Metrics}
We evaluate on two automatic evaluation metrics: SentencePiece BLEU \citep{papineni-etal-2002-bleu}, using the FLORES-200 tokenizer, and chrF++ 
\citep{popovic-2017-chrf}\footnote{We conduct evaluation using the \href{https://github.com/mjpost/sacrebleu}{sacreBLEU library.}}. We choose both metrics for consistency with the NLLB-200 evaluation methodology \citep{nllbteam2022language}. We avoid using model-based automatic evaluation metrics such as COMET \citep{rei-etal-2020-comet} since proper evaluation of very low-resource languages is critical to our methodology.

\subsection{Baselines}
We benchmark Claude performance against the strongest multilingual baselines, NLLB-200 and Google Translate. 
Prior work has examined the performance of various LLMs on the FLORES dataset \cite{zhu2023multilingual, robinson-etal-2023-chatgpt}. Previously, the NLLB-200 multilingual translation model \cite{nllbteam2022language} has outperformed existing LLMs, especially on low-resource languages. Meanwhile, commercial translation systems (e.g Google Translate) have outperformed NLLB, especially on medium to high resource languages \cite{zhu2023multilingual}. We benchmark against the best NLLB model (NLLB-54B), and against the public-facing Google Translate API.

\subsection{Claude Translation Methodology}
To generate translations from Claude, we use the model Claude 3 Opus. All experiments were run between March 2024 and April 2024. 

\subsubsection{Prompt Tuning}
Prior work has optimized LLM prompts for machine translation. Following the findings of \cite{zhang2023prompting}, we use a tuneable number of in-context sentence exemplars within the prompt, drawn from the \texttt{dev} split of the FLORES-200 dataset. We also introduce a new prompt used to ``batch'' translations from the same document together, leading to translation quality improvement and reduced API cost. The exact prompts are provided Appendix \ref{sec:appendix:example-prompts}, with examples shown in Table \ref{tab:sentence-prompt-example} and Table \ref{tab:document-prompt-example}. We use 8 in-context exemplars for sentence-level prompts and 1 in-context exemplar for document-level prompts.

\subsubsection{Temperature Tuning}
The Claude API allows \texttt{temperature} as input, ranging from 0 to 1. Based on tuning experiments, we set \texttt{temperature=0.7} for each evaluation.


\section{Results}
We begin by outlining the empirical performance of Claude on the FLORES datasets, and subsequently compare the results to unseen datasets such as the BBC News datasets and the Maltese speech dataset.
\label{sec:results}

\paragraph{When translating into English, Claude surpasses the baselines on the majority of the considered language pairs in the FLORES-200 dataset.} We showcase the full results on all 36 language pairs in Table \ref{tab:flores-translation-bleu} and Table \ref{tab:flores-translation-chrf}.

The chrF++ score of Claude exceeds the baselines for 58\% of language pairs in the \texttt{xxx->eng} direction and for 11\% of the evaluated language pairs in the \texttt{eng->xxx} direction. However, these results should not be taken as definitive evidence of Claude's translation ability, as we will show that Claude demonstrates signs of data contamination on the FLORES-200 benchmark.

\begin{figure*}[t]
    \centering
    \begin{subfigure}[b]{0.48\textwidth}
        \centering
        \includegraphics[width=\textwidth]{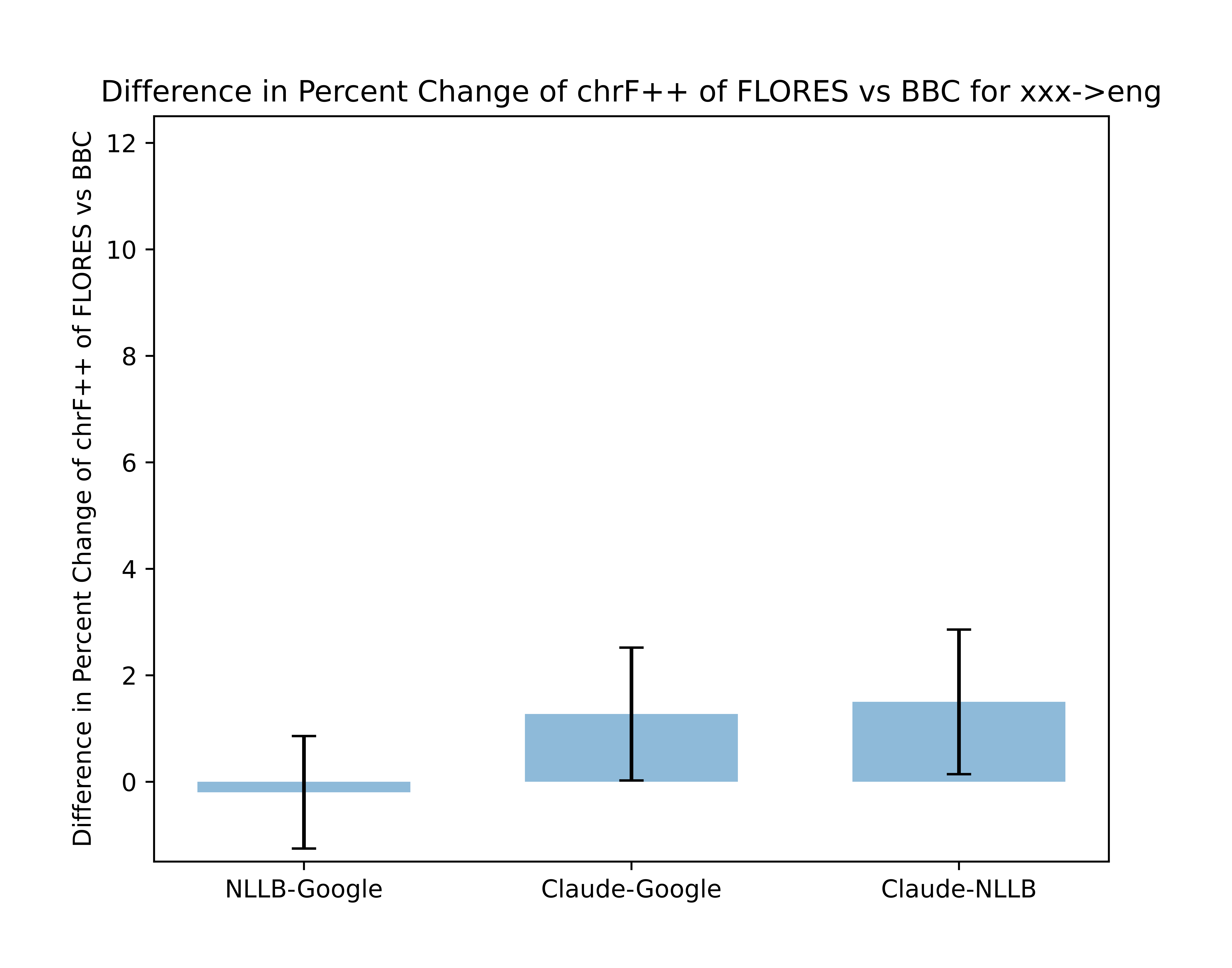}
        \caption{Comparison for the \texttt{xxx->eng} direction.}
        \label{fig:flores-bbc-xxx-eng}
    \end{subfigure}
    \hfill
    \begin{subfigure}[b]{0.48\textwidth}
        \centering
        \includegraphics[width=\textwidth]{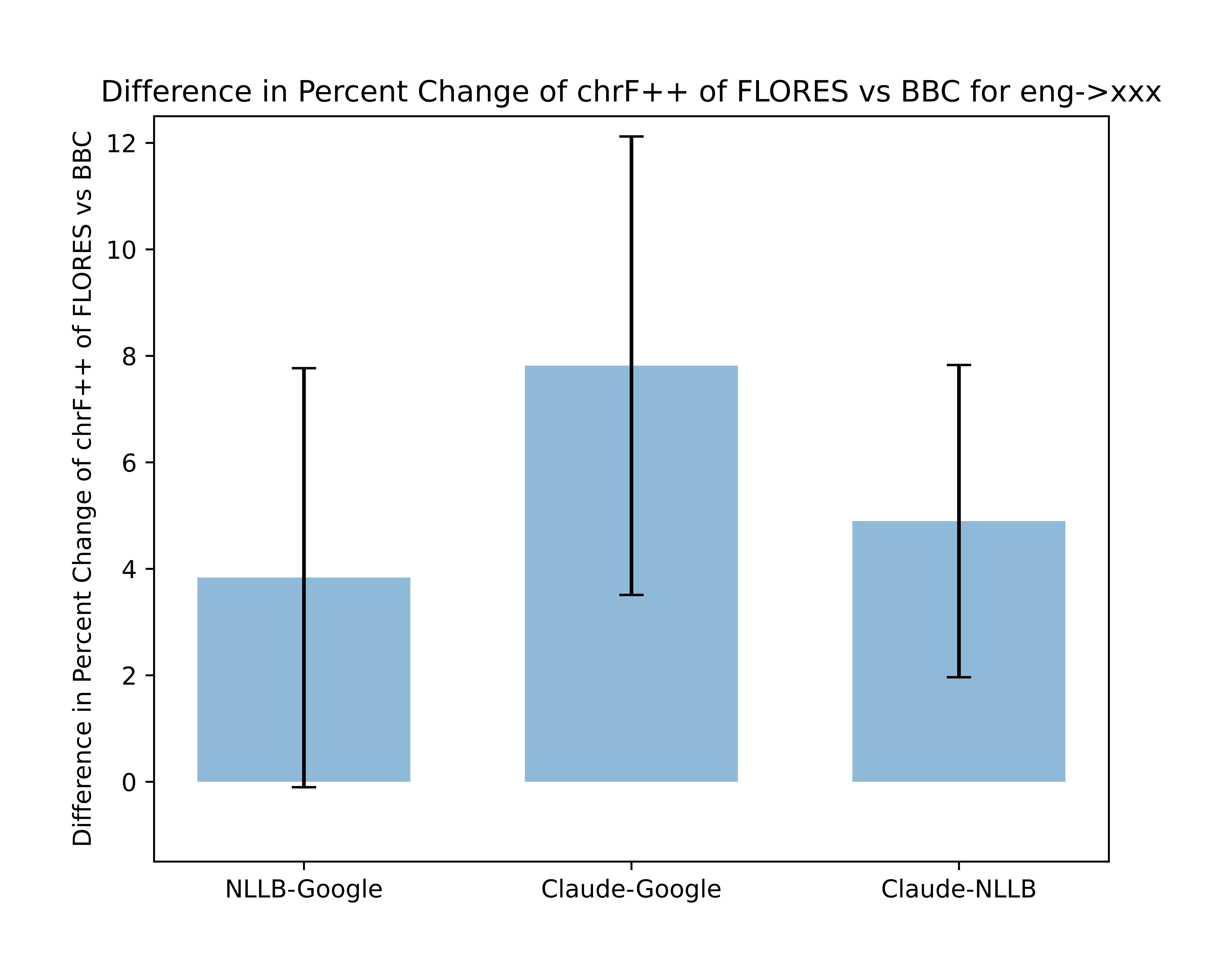}
        \caption{Comparison for the \texttt{eng->xxx} direction.}
        \label{fig:flores-bbc-eng-xxx}
    \end{subfigure}
    \caption{Comparison of relative performance of between FLORES and BBC datasets.}
    \label{fig:flores-bbc-comparison}
\end{figure*}

\paragraph{On unseen BBC datasets, Claude beats the SOTA on a number of both low and high-resource languages when translating into English.}
We evaluated all 36 language pairs in the \verb|eng->xxx| and \verb|xxx->eng| translation directions. We provide the complete results in Table \ref{tab:bbc-translation-bleu} and Table \ref{tab:bbc-translation-chrf}.

In summary, Claude surpasses the state-of-the-art on Bengali-English, French-English, Kyrgyz-English, Korean-English, Nepali-English, Russian-English, Ukrainian-English, and Yoruba-English, English-Korean, and English-Thai, constituting 11 translation directions out of the 72 total. These account for 25\% of language pairs in the \texttt{xxx->eng} direction and 5.5\% of language pairs in the \texttt{eng->xxx} direction. The languages cover a wide array of scripts and language families, indicating that Claude-based translation is not necessarily biased towards languages similar to English. In Section \ref{sec:knowledge-distillation}, we will show that all of these languages may be amenable to knowledge distillation techniques in order to advance the state-of-the-art.

\paragraph{Claude shows signs of data contamination on the FLORES-200 dataset in both translation directions.}
It is likely that Claude has seen the FLORES data during its training, but it remains unclear whether this measurably affects Claude's performance on the benchmark. We investigate this question by comparing the results on FLORES versus BBC News. Because the FLORES and BBC datasets may vary in difficulty and quality, we cannot directly compare the raw chrF++ scores of each model across the datasets. However, we expect that some model has no dataset contamination relative to another model if the relative performance between the models is similar between the dataset in question and unseen data.

In Figure \ref{fig:flores-bbc-comparison}, we visualize this difference. In the \texttt{xxx->eng} direction, we observe that Google and NLLB have very similar performance across the FLORES and BBC datasets, indicating little-to-no contamination of either dataset for either model. 

However, we observe substantial increase in performance of Claude on FLORES compared to BBC relative to either Google or NLLB, which suggests that Claude has overfit the FLORES dataset, with its performance overrepresented by 1-2 percentage points. This analysis calls into question the validity of evaluating Claude on FLORES.

In the \texttt{eng->xxx} direction, we observe somewhat more complicated behavior. As before, Claude also performs relatively worse on BBC than the other models, and to a significantly larger extent. Thus, LLMs may be more prone to contamination in the \texttt{eng->xxx} direction. However, the NLLB-Google column also suggests dataset contamination, such that either NLLB has overfit FLORES or Google has overfit BBC. We reject the first possibility since it is known that NLLB has not been trained on FLORES \citep{nllbteam2022language}; therefore, Google may be biased toward the BBC benchmark in the \verb|eng->xxx| direction.

We offer a possible explanation of these findings: when BBC authors write translations of existing English articles into other languages, they might use Google Translate to provide candidate translations of the article before they edit the writing to improve fluency. This procedure introduces a bias towards Google-like translations in the \texttt{eng->xxx} direction. Thus, we avoid evaluating on Google in the \texttt{eng->xxx} BBC direction for the remainder of the analysis.

\paragraph{Claude's translation ability is better when translating into English rather than out of English.}
\citet{zhu2023multilingual} observe that LLMs are better at translating into English, and we observe the same effect. Claude improves over the SOTA in a much smaller percentage of language pairs in the \texttt{eng->xxx} direction. 


According to Table \ref{tab:bbc-translation-chrf}, Claude exceeds NLLB-54B on 56\% of language pairs in \texttt{xxx->eng} translation but only 33\% of pairs in \texttt{eng->xxx} translation. Further, the mean improvement of Claude over NLLB is 0.81\% in \texttt{xxx->eng} translation but -3.05\% in \texttt{eng->xxx} translation. Thus, Claude has impressive translation into English but still struggles with out-of-English translation.

\begin{figure*}[t]
    \centering
        \begin{subfigure}[b]{0.48\textwidth}
        \centering
        \includegraphics[width=\textwidth]{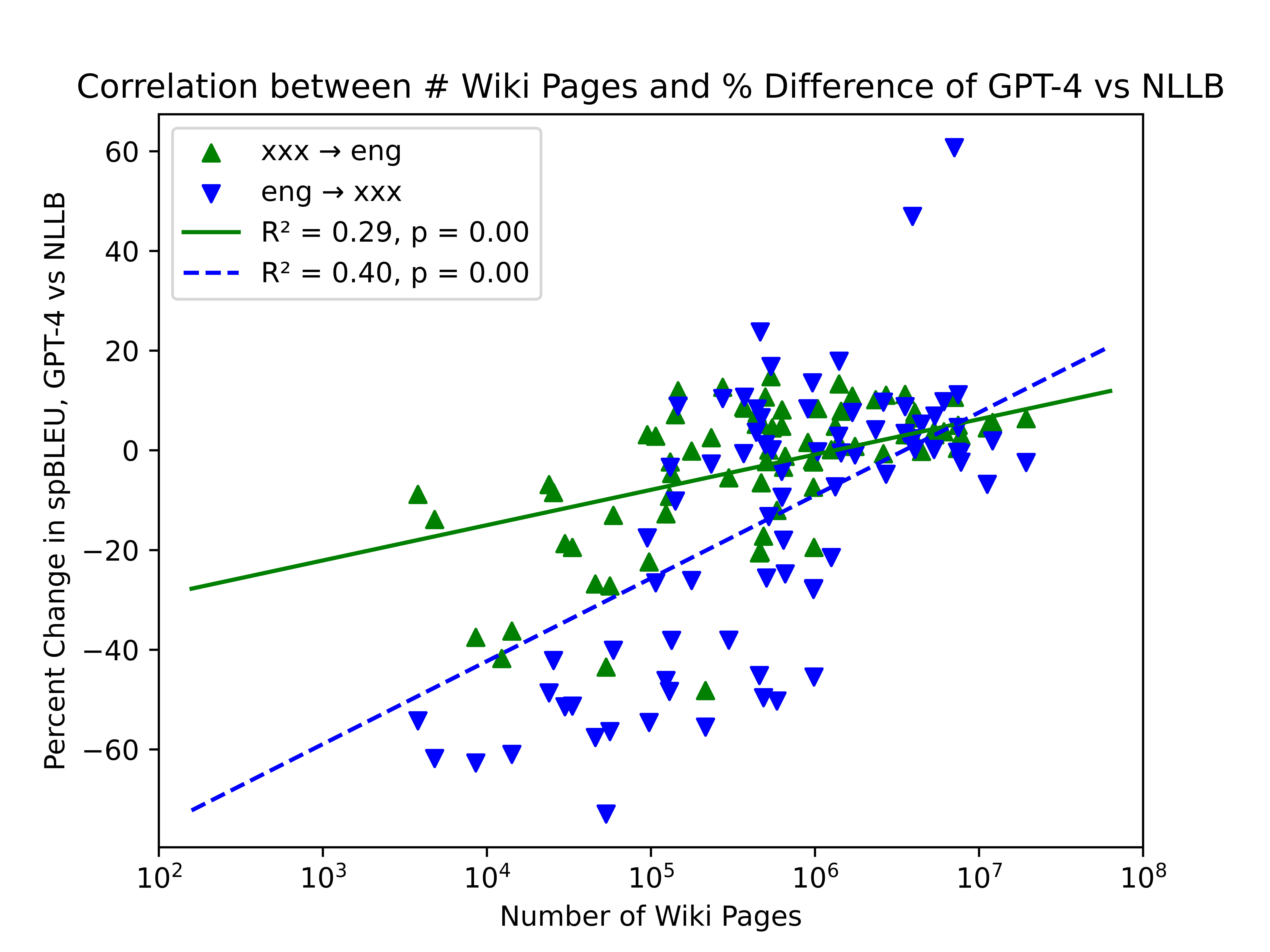}
        \caption{Correlation between number of Wikipedia pages and GPT-4 relative performance compared to NLLB-1.3B.}
        \label{fig:gpt-4-correlation}
    \end{subfigure}
    \hfill
    \begin{subfigure}[b]{0.48\textwidth}
    \centering
        \includegraphics[width=\textwidth]{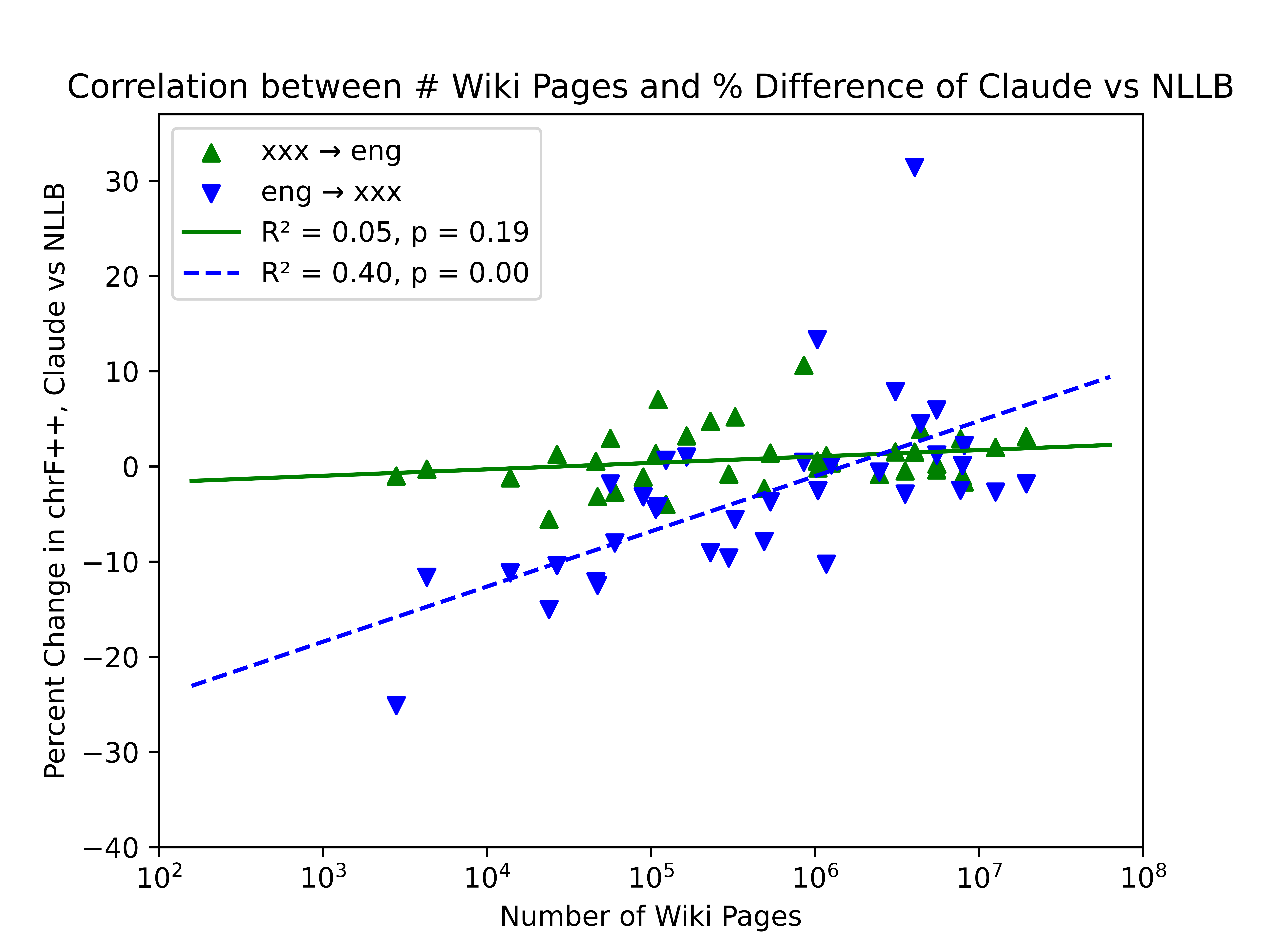}

        \caption{Correlation between number of Wikipedia pages and Claude relative performance compared to NLLB-54B.}
        \label{fig:claude-correlation}
    \end{subfigure}
        \caption{Comparison of performance correlations between Claude and GPT-4, relative to NLLB models. Figure \ref{fig:claude-correlation} uses our data and Figure \ref{fig:gpt-4-correlation} is generated from data from \citet{zhu2023multilingual}.}
    \label{fig:wiki-pages-correlation-comparison}
\end{figure*}

\paragraph{Unlike other evaluated LLMs, Claude has remarkable \textit{resource efficiency} when translating into English.} Previous works \citep{zhu2023multilingual, stap-araabi-2023-chatgpt, robinson-etal-2023-chatgpt} have found that LLMs are more competitive with multilingual NMT models (like NLLB) when translating high-resource (rather than low-resource) language pairs. We dig deeper into this claim and evaluate the extent to which it is true for Claude-based translation.

By \textit{resource efficiency}, we mean the extent to which the performance of a multilingual translation engine depends on the resource level (e.g. high, low, very low) of the language pair. Previous work \cite{zhu2023multilingual} has shown that LLMs have inferior ability to translate low-resource languages relative to NMT models like NLLB-54B, indicating poor relative resource efficiency\footnote{Note that \citet{zhu2023multilingual} claims that LLMs have good resource efficiency after finding that XGLM 7.5B can translate some \textit{unresourced} languages. However, the same work finds that LLMs lag behind specialized NMT models when translating low-resource languages.}.

To quantify the resource efficiency of a multilingual machine translation system A \emph{relative to a baseline system B}, we perform a linear regression on the performance of A relative to B on a language pair against an independent variable measuring the resource level of the language pair. Following \cite{robinson-etal-2023-chatgpt}, we use the number of Wikipedia articles\footnote{\href{https://web.archive.org/web/20230723062438/https://en.wikipedia.org/wiki/List_of_Wikipedias}{https://en.wikipedia.org/wiki/List\_of\_Wikipedias}} as our independent variable, and measure the percent difference between our LLM of interest and the NLLB baseline model. We then run a t-test to assess the significance that the slope coefficient is nonzero\footnote{We use the  Wald Test with t-distribution of the test statistic \citep{wald1943tests}, which is default in \texttt{scipy}.}, using a significance cutoff of 0.05. A positive slope with significant p-value indicates that the resource-level positively predicts the translation quality of the LLM relative to NLLB, so the LLM has low resource efficiency. A slope close to zero with non-significant p-value indicates that the resource efficiency is close to NLLB. Finally, a significantly negative slope indicates that the LLM is \textit{more} resource efficient than the supervised baseline.

Using this setup, we verify that 8 LLMs evaluated in \citet{zhu2023multilingual}, including LLAMA and GPT-4, all exhibit significant correlation on resource level with respect to comparative performance to NLLB, which indicates that NLLB is more resource efficient than these other LLMs. In Figure \ref{fig:gpt-4-correlation}, we plot the correlation for GPT-4 (separated by translation direction), where the data is collected from \citet{zhu2023multilingual}. In both translation directions, the GPT-4 model has significant correlation with respect to resource level on performance relative to NLLB. We provide similar plots for 7 other LLMs of interest in Appendix \ref{fig:correlation-plots}. 

However, there is one outlier: Claude. In Figure \ref{fig:claude-correlation}, we show that in \texttt{xxx->eng} translation, Claude has comparable resource efficiency to NLLB. Claude may be the first LLM to demonstrate resource efficiency in machine translation versus strong NMT baselines. Thus, among current LLMs, Claude shows particular promise as a low-resource translator.

\paragraph{Claude outperforms NLLB on the IWSLT 2024 Maltese-English Shared Task dataset.}

Our next dataset comes from the IWSLT 2024 Shared Task in low-resource machine translation. We consider the performance of Claude on the development split of the unseen, parallel MASRI-HEADSET dataset in the Maltese speech domain \citep{hernandez-mena-etal-2020-masri}\footnote{Although the dataset was created before the training cutoff of Claude, it requires access to be granted in order to be downloaded. Furthermore, translations were not created until 2024.}.



\begin{figure}[t]
    \centering
    \includegraphics[width=.45\textwidth]{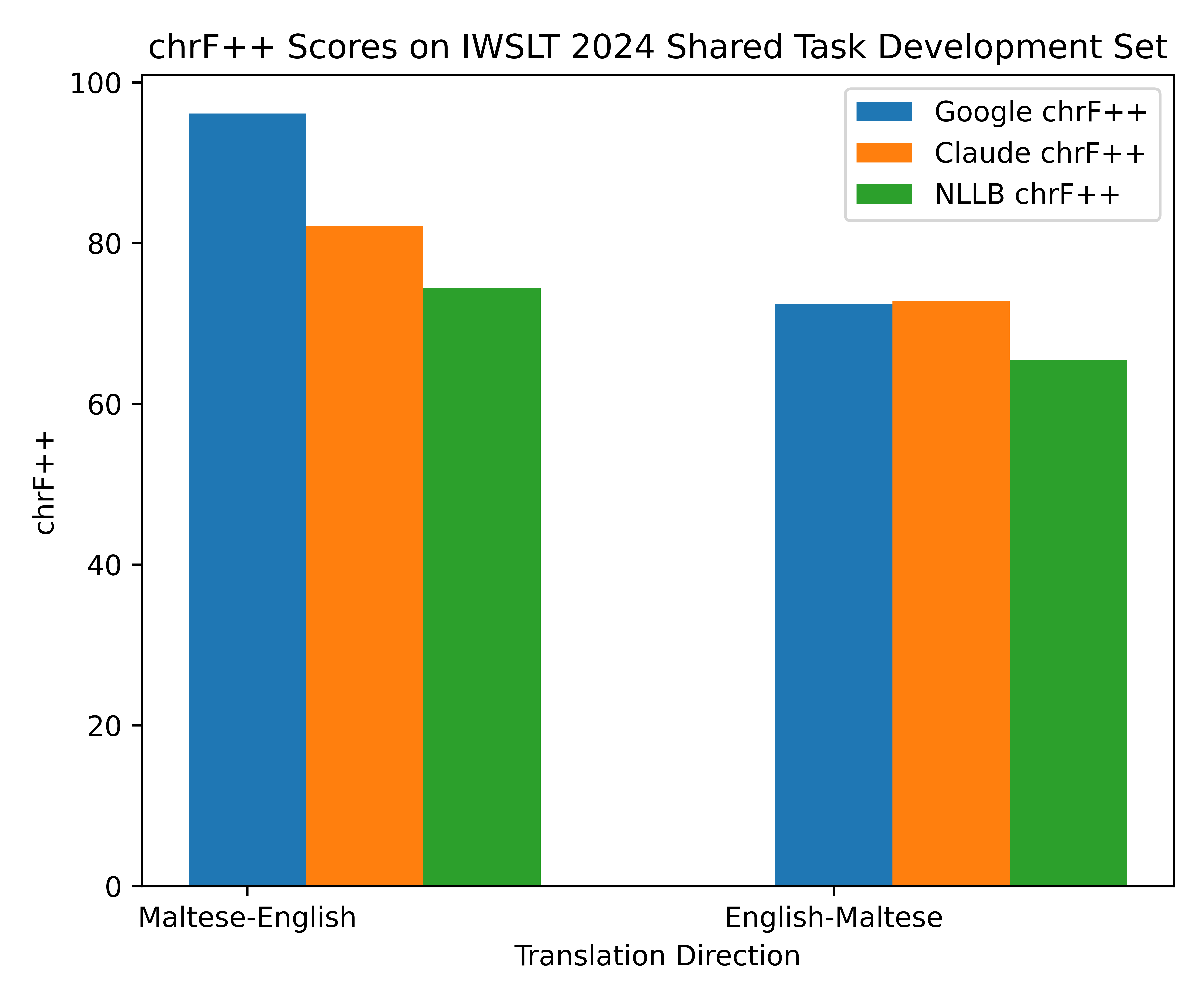}
    \caption{Translation performance, in chrF++, of the models on the development set of the IWSLT 2024 Maltese-English shared task.}
    \label{fig:maltese-results}
\end{figure}

In Figure \ref{fig:maltese-results}, we display the performance of Google, NLLB, and Claude on the MASRI-HEADSET dataset. 
In the Maltese-English direction, we observe unusual translation behavior: Google demonstrates nearly perfect BLEU and chrF++. Even a perfect translator usually should not have perfect scores against reference translations due to the intrinsic variations of natural language. The results are drastically different compared to the English-Maltese translation direction. One possible explanation is that the dataset translations were automatically translated from Maltese into English by Google Translate and then post-edited. 


Nevertheless, Claude outperforms NLLB in both the Maltese-English and English-Maltese directions. Thus, Claude demonstrates robust translation ability across multiple domains.


\section{Knowledge Distillation}
\label{sec:knowledge-distillation}
Although LLMs may achieve state-of-the-art results in certain translation directions, the cost, time, and energy use of computational inference limits their applicability as translators. For example, a system such as Google Translate needs cheap inference to support the billions of words translated on a daily basis \citep{Turovsky2016}. The task of compressing model performance into smaller models is known as knowledge distillation, and has been studied both in the broader deep learning literature \cite{hinton2015distilling}, and for machine translation \cite{kim2016sequencelevel}. In this section, we devise LLM-based knowledge distillation methods. 

\subsection{Yoruba}
Spoken primarily in the West African countries Nigeria, Benin, and Togo, Yoruba has over 44 million native speakers \cite{ethnologue2024yoruba}. However, Yoruba is low-resource and existing NMT models are low-quality. According to our results in Table \ref{tab:bbc-translation-chrf}, Claude may be able to translate Yoruba-English better than Google and NLLB-54B, showing promise as a potential case study for knowledge distillation. 


\subsection{Sequence-KD}
Sequence-KD \citep{kim2016sequencelevel} is a knowledge distillation method which involves translating the source side of the teacher's training corpus via beam search, and then training a student model on the translated corpus. Here, we generate translations of a monolingual Yoruba corpus using the LLM. Following the methodology of \citet{dabre-fujita-2020-combining}, we fine-tune a strong base model on the small forward-translated corpus.


\subsection{Document Translation}
We translate a new monolingual corpus of Yoruba news articles crawled from \href{https://www.bbc.com/yoruba}{bbc.com/yoruba}, ignoring all articles written after June 1, 2023 to ensure no overlap with the test corpus. To reduce number of inferences requested from Claude, and reuse in-context-exemplars across multiple sentences, we translate each article with one API request. Previous work has shown that LLMs have impressive document-level translation ability \cite{karpinska-iyyer-2023-large,wang-etal-2023-document-level}, although none have studied sentence-level document translation, where the aligned sentence translations must be recovered from the original document and the translated document. To achieve sentence-aligned document translation, we use the prompts described in Appendix \ref{sec:appendix:claude-prompts}. An example prompt is shown in Table \ref{tab:document-prompt-example}.

From the monolingual crawled corpus, we forward-translate 53193 sentences with NLLB-54B to create the base distillation model. Next, to create the Claude and Google distillation datasets, we forward-translate a selection of 431 Yoruba news documents totaling 7996 Yoruba sentences. The total API cost of parallel corpus generation using Claude 3 Opus was \$46.06. 

\begin{table*}
\centering
\small
\begin{tabular}{|c|c|c|c|c|c|c|c|c|c|}
\hline
 & \multicolumn{4}{c|}{Baselines} & \multicolumn{3}{c|}{Distillation Models} \\
\hline
Metric & \verb|google| & \verb|nllb-54B| & \verb|claude_sent| &\verb|claude_doc| & \verb|base_distill| & \verb|google_distill| & \verb|claude_distill|\\
\hline
spBLEU & 21.09 & 22.51 & 21.15 & \textbf{26.17} & \textbf{22.82} & 20.81 &22.61\\
chrF++ & 41.99 & 43.26 & 43.78 & \textbf{47.06} & 43.23 & 42.29 &\textbf{44.15}\\
\hline
\end{tabular}
\caption{spBLEU, chrF++ scores on the Yoruba-English BBC News dataset. Bolded results are best in each category.}
\label{tab:kd-results}
\end{table*}

\subsection{Experiments}
Our approach is to train a strong base model by distilling NLLB-54B Yoruba-English into a smaller model. We use the NLLB-54B distillation dataset, as well as the train split of the MENYO-20k dataset \citep{adelani-etal-2021-effect}, which contains human-labeled Yoruba-English parallel sentences across multiple domains. Then, we fine-tune NLLB-1.3B\footnote{Model card: \href{https://huggingface.co/facebook/nllb-200-distilled-1.3B}{facebook/nllb-200-distilled-1.3B}} on a shuffling of these two datasets to create our base distillation model \verb|base_distill|.
To train the distillation models \verb|claude_distill| and \verb|google_distill|, we fine-tune \verb|base_distill| on the respective synthetic forward-translated corpus, combined with a random sampling from the NLLB-54B distillation corpus and MENYO-20k train corpus in equal ratio 1:1:1.

The model \verb|base_nllb_distill| is trained until validation loss is minimized, which took 1 epoch. The models \verb|claude_distill| and \verb|google_distill| are trained until spBLEU on the validation set is maximized, which took 2 epochs. We used an AdamW optimizer \cite{loshchilov2019decoupled} with an initial learning rate of 5e-5 for \verb|base_distill| and 3e-5 for \verb|claude_distill| and \verb|google_distill|. Each model is trained with batch size 6. To compute translations from each model, we use beam search with 5 beams and restrict repetition of n-grams of 3 or larger.

\subsection{Distillation Results}
The results of the experiments are displayed in Table \ref{tab:kd-results}. Under the ``Baselines'' category, the columns \verb|claude_sent| and \verb|claude_doc| refer to translating with Claude using the sentence-level prompt or the document-level prompt (see Appendix \ref{sec:appendix:claude-prompts}).

The best model by far is \verb|claude_doc|, surpassing all other models by over 3 spBLEU and chrF++. These results suggest that document-level context can substantially improve translation quality of Claude.

We observe that our model, \verb|claude_distill|, attains higher chrF++ than all other non-Claude baselines, including Google Translate and NLLB-54B. The model also has considerably better performance than \verb|google_distill|, which demonstrates the importance of data quality when augmenting a training corpus with synthetic data. Thus, by distilling on a relatively small dataset with Claude, we are able to match or exceed the performance of \verb|claude_sent|, and construct a small model that outperforms the baselines.

\section{Limitations}
Due to a limited budget for the Claude API, our per-language dataset size was constrained. Moreover, all evaluated language pairs involved English as the source or target. Finally, since all unseen data comes from BBC News, our evaluation strategy would not directly apply to languages that BBC News does not support. Since evaluating the translation performance of LLMs on published MT benchmarks can be problematic due to data contamination, an important question remains on how to evaluate Claude (and other closed-source LLMs) on a broader set of languages.

\section{Conclusions}
Our results point toward a future era of LLM-powered machine translation. Although we find that Claude shows signs of data contamination on FLORES-200, we also evaluate Claude on unseen datasets and find that Claude 3 Opus outperforms NLLB-54B on 44\% of language pairs and Google Translate on 22\%. Unlike prior LLM models, the spBLEU and chrF++ scores of Claude remain competitive, or even exceed, the baseline models on high, low, and very-low resource language pairs. In fact, among 8 other LLMs, we show that Claude uniquely demonstrates a \textit{resource efficiency} comparable to NLLB-54B.  Finally, in section \ref{sec:knowledge-distillation}, we show that state-of-the-art results from LLMs can be distilled into inexpensive machine translation models and we create a simple system that beats baselines on Yoruba-English for BBC News articles.

Due to the increasing capabilities of LLMs as models scale in size and efficiency, we expect that (potentially closed-source) LLMs will surpass the state-of-the-art in more and more language pairs. Our work demonstrates that these advancements can be harnessed by the MT community to improve under-resourced language pairs.

This work opens many interesting avenues of future research. Our evaluations are limited to English-centric translation, but our methods (and automatic dataset construction) should apply to any language pair. \citet{zhu2023multilingual} has shown that LLMs struggle with non-English-centric machine translation - can we use the mined BBC dataset methods to evaluate new LLMs and prompting techniques on non-English-centric translation?

% ...

\bibliographystyle{acl_natbib}
\bibliography{custom,anthology}

\clearpage
\onecolumn
\appendix
\section{Appendix}
\subsection{Language Table}
\small
\begin{longtable}
{p{0.2\textwidth}p{0.1\textwidth}p{0.08\textwidth}p{0.09\textwidth}p{0.15\textwidth}p{0.16\textwidth}p{0.1\textwidth}}
\hline
Language & Code & Script & ISO 639-1 & Family & Subgrouping & Res. \\
\hline
\endhead
\hline
\endfoot
Amharic & amh\_Ethi & Ge'ez & am & Afro-Asiatic & Semitic & Low\\
Modern Standard Arabic & arb\_Arab & Arabic & ar & Afro-Asiatic & Semitic & High\\
Azerbaijani & azj\_Latn & Latin & az & Turkic & Common Turkic & Very Low\\
Bengali & ben\_Beng & Bengali & bn & Indo-European & Indo-Aryan & High\\
English & eng\_Latn & Latin & en & Indo-European & Germanic & High\\
French & fra\_Latn & Latin & fr & Indo-European & Italic & High\\
Western Central Oromo & gaz\_Latn & Latin & om & Afro-Asiatic & Cushitic & Low\\
Gujarati & guj\_Gujr & Gujarati & gu & Indo-European & Indo-Aryan & Low\\
Hausa & hau\_Latn & Latin & ha & Afro-Asiatic & Chadic & Low\\
Hindi & hin\_Deva & Devanagari & hi & Indo-European & Indo-Aryan & High\\
Igbo & ibo\_Latn & Latin & ig & Atlantic-Congo & Benue-Congo & Low\\
Indonesian & ind\_Latn & Latin & id & Austronesian & Malayo-Polynesian & High\\
Japanese & jpn\_Jpan & Japanese & ja & Japonic & Japanese & High\\
Kinyarwanda & kin\_Latn & Latin & rw & Atlantic-Congo & Benue-Congo & Low\\
Kyrgyz & kir\_Cyrl & Cyrillic & ky & Turkic & Common Turkic & Low\\
Korean & kor\_Hang & Hangul & ko & Koreanic & Korean & High\\
Marathi & mar\_Deva & Devanagari & mr & Indo-European & Indo-Aryan & Low\\
Burmese & mya\_Mymr & Myanmar & my & Sino-Tibetan & Burmese-Lolo & Low\\
Nepali & npi\_Deva & Devanagari & ne & Indo-European & Indo-Aryan & Very Low\\
Southern Pashto & pbt\_Arab & Arabic & ps & Indo-European & Iranian & Very Low\\
Persian & pes\_Arab & Arabic & fa & Indo-European & Iranian & High\\
Portuguese & por\_Latn & Latin & pt & Indo-European & Italic & High\\
Russian & rus\_Cyrl & Cyrillic & ru & Indo-European & Balto-Slavic & High\\
Sinhala & sin\_Sinh & Sinhala & si & Indo-European & Indo-Aryan & Low\\
Somali & som\_Latn & Latin & so & Afro-Asiatic & Cushitic & Very Low\\
Spanish & spa\_Latn & Latin & es & Indo-European & Italic & High\\
Swahili & swh\_Latn & Latin & sw & Atlantic-Congo & Benue-Congo & Low\\
Tamil & tam\_Taml & Tamil & ta & Dravidian & South Dravidian & Low\\
Telegu & tel\_Telu & Telegu & te & Dravidian & South Dravidian & Low\\
Thai & tha\_Thai & Thai & th & Kra-Dai & Tai & Low\\
Tigrinya & tir\_Ethi & Ge'ez & ti & Afro-Asiatic & Semitic & Low\\
Turkish & tur\_Latn & Latin & tr & Turkic & Common Turkic & High\\
Ukrainian & ukr\_Cyrl & Cyrillic & uk & Indo-European & Balto-Slavic & High\\
Urdu & urd\_Arab & Arabic & ur & Indo-European & Indo-Aryan & Low\\
Northern Uzbek & uzn\_Latn & Latin & uz & Turkic & Common Turkic & High\\
Vietnamese & vie\_Latn & Latin & vi & Austroasiatic & Viet-Muong & High\\
Yoruba & yor\_Latn & Latin & yo & Atlantic-Congo & Benue-Congo & Low\\
\hline
\caption{The list of 37 languages used for experimentation. All experiments are on English-centric translation, giving a total of 72 translation directions.}
\label{tab:language-table}
\end{longtable}

\twocolumn
\subsection{Claude Prompts}
\label{sec:appendix:claude-prompts}
\begin{table*}[]
    \centering
    \begin{tabular}{l|p{14cm}}
    \hline
    Prompt & French: Au fil des siècles, les gens ont soigneusement construit des terrasses sur le paysage accidenté et escarpé jusqu'aux falaises qui surplombent la mer. English: Over the centuries, people have carefully built terraces on the rugged, steep landscape right up to the cliffs that overlook the sea. French: 800 miles du réseau de pipelines Trans-Alaska ont été fermés suite à un déversement de milliers de barils de pétrole brut au sud de Fairbanks, en Alaska. English: 800 miles of the Trans-Alaska Pipeline System were closed down following a spill of thousands of barrels of crude oil south of Fairbanks, Alaska.\\
    & French:  On pense qu'il se présentera à la présidence en 2016. English:\\
    \hline

    Output & He is thought to be running for president in 2016.\\
    \hline
    \end{tabular}
    \caption{An example of a sentence-level translation prompt in a French-English translation task on the FLORES-200 dataset, with 2 in-context exemplars originated from the dev split of FLORES-200. Note that in-context exemplars are not separated by newlines, according to \cite{zhang2023prompting}.}
    \label{tab:sentence-prompt-example}
\end{table*}

\label{sec:appendix:example-prompts}
In this subsection, we formally specify the sentence-level prompt used in Section \ref{sec:results} and document-level prompt used in Section \ref{sec:knowledge-distillation}, and provide examples of both.
\\
\\
\textbf{Sentence prompt} We generate our sentence-level prompts in the following format:
\begin{verbatim}
    {source}: {X_1} {target}: {Y_1}
    ... {source}: {X_n} {target}: {Y_n}\n
    {source}: {X'} {target}: 
\end{verbatim}

where \verb|source| is the source language (e.g \texttt{English}, \texttt{Spanish}, etc.), \verb|target| is the target language, \verb|X_i| is the i'th source sentence, \verb|Y_i| is the i'th target sentence (the gold translation of \verb|X_i|), and \verb|X'| is the desired sentence to translate. Here, $n$ is a hyperparameter specifying the number of in-context exemplars. This prompt is taken from \citet{zhang2023prompting}. An example prompt on a French-English translation task is provided in Table \ref{tab:sentence-prompt-example}.\\\\
\textbf{Document prompt} We define a \textit{single exemplar} of documents \verb|X| and \verb|Y|:
\begin{verbatim}
    {source}:\n
    1. {X[1]}\n
    2. {X[2]}\n
    .
    .
    .
    {i}. {X[i]}\n
    \n
    Line-by-line {target} translations:\n
    1. {Y[1]}\n
    2. {Y[2]}\n
    ...
    {i}. {Y[i]}
\end{verbatim}
where \verb|X| is a document with $i$ sentences in the language given by \verb|source| and \verb|Y| is the \textit{gold translation} document of $i$ sentences in the language given by \verb|target| (where each sentence \verb|Y[j]| is a translation of \verb|X[j]|).

Then, we define the \textit{query prompt}:
\begin{verbatim}
    {source}:\n
    1. {X'[1]}\n
    2. {X'[2]}\n
    .
    .
    .
    {k}. {X'[k]}\n
    \n
    Line-by-line {target} translations:\n
\end{verbatim}

where \verb|X'| is the desired document with \verb|k| sentences for which to generate line-by-line translations.

Then, the entire sentence-aligned document-level prompt is composed of \verb|n| single exemplars followed by a query prompt, joined by double newlines. See Table \ref{tab:document-prompt-example} for an example of a document-level prompt and the respective Claude 3 output.

\begin{table*}[]
    \centering
    \begin{tabular}{l|p{14cm}}
    \hline
    Prompt & French:\\
            & 1. La manifestation a commencé vers 11 h heure locale (UTC+1) sur Whitehall, en face de l'entrée gardée par la police de Downing Street, la résidence officielle du Premier ministre.\\
            & 2. À Whitehall, peu après 11 h 00, des manifestants ont bloqué la circulation des véhicules en direction du nord.\\
            & 3. À 11 h 20, la police a demandé aux manifestants de retourner sur le trottoir, déclarant qu’ils devaient trouver un équilibre entre le droit de manifester et l’augmentation de la circulation.\\
            & \\
            & Line-by-line English translations:\\
            & 1. The protest started around 11:00 local time (UTC+1) on Whitehall opposite the police-guarded entrance to Downing Street, the Prime Minister's official residence.\\
            & 2. Just after 11:00, protesters blocked traffic on the northbound carriage in Whitehall.\\
            & 3. At 11:20, the police asked the protesters to move back on to the pavement, stating that they needed to balance the right to protest with the traffic building up.\\
            & \\
            & French:\\
            & 1. Le Haut Karabakh déserté révèle les séquelles d'une défaite arménienne fulgurante \\
            & 2. Dans un avant-poste arménien situé dans les montagnes du Haut-Karabakh, une marmite à moitié pleine se trouvait à côté d’une assiette de nourriture à moitié mangée.\\
            & 3. Il y avait une cigarette à moitié fumée et un morceau de pain.\\
            & 4. Dans un autre avant-poste plus petit, plus loin le long de l'ancienne ligne de front, un journal de bord arménien était abandonné dans l'herbe.\\
            & \\
            & Line-by-line English translations:\\
    \hline

    Output & 1. The deserted Nagorno-Karabakh reveals the aftermath of a lightning Armenian defeat\\
        & 2. In an Armenian outpost located in the mountains of Nagorno-Karabakh, a half-full pot was next to a half-eaten plate of food. \\
        & 3. There was a half-smoked cigarette and a piece of bread.\\
        & 4. In another smaller outpost further along the former front line, an Armenian logbook was abandoned in the grass.\\
    \hline
    \end{tabular}
    \caption{An example of a sentence-aligned document-level translation prompt in a French-English translation task, with one in-context exemplar. The in-context exemplar comes from the FLORES-200 dataset and the translation task comes from \href{https://www.bbc.com/afrique/articles/c0kxzprpnqqo}{https://www.bbc.com/afrique/articles/c0kxzprpnqqo}. }
    \label{tab:document-prompt-example}
\end{table*}

\newpage
\subsection{Creation of Mined BBC Datasets}
\label{sec:appendix:bbc-mined}
Our dataset creation procedure involves six main steps:
\begin{enumerate}
    \item \textbf{Monolingual BBC page collection} We begin by accessing the Web Archive API\footnote{\href{https://archive.org/developers/wayback-cdx-server.html}{https://archive.org/developers/wayback-cdx-server.html}} in order to find monolingual BBC webpages in the source language.
    \item \textbf{Google Reverse Image Search proposals} We use Google Reverse Image Search to generate candidate translation pages for each source article. In some cases, multiple candidate pages or no candidate pages are proposed for each article.
    \item \textbf{Sentence splitting} We create a multilingual sentence splitter and split the source and target articles into sentences.
    \item \textbf{Per-document sentence alignment} We use the Facebook LASER library\footnote{\href{https://github.com/facebookresearch/LASER}{https://github.com/facebookresearch/LASER}} to align sentences between candidate source and target articles. For each source article, we use the \texttt{intersection} mining technique to find candidate sentences alignments across a given target article. If multiple target documents were proposed, we then select the document and sentence translations which maximize the product of LASER score across the given source document.
    \item \textbf{Date filtration} We filter all source-target candidate translations where the English BBC document has date beyond September 1, 2023\footnote{As of the publication date of the paper, this date is the beyond the training cutoff for Claude.} to ensure that the BBC articles are unseen by Claude.
    \item \textbf{LASER score filtration} We filter all sentence alignments with LASER score less than 1.03. Then, we collect the sentence alignments into a dataset coming from source-target documents by the largest LASER score until we accumualate more than 100 sentences.
\end{enumerate}

Note that due to the candidate article proposal step (Step 4), this procedure is very computationally inexpensive, and is bottlenecked only by HTTPS response time. Thus, we can use the technique to generate larger datasets, if required. We can also modify the date filtration step to collect unseen data for any given LLM with a training cutoff date sufficiently far in the past.

Finally, we briefly note that the sentence alignment procedure may create a dataset that bias towards NLLB, since NLLB has been trained on LASER-mined parallel datasets \citep{nllbteam2022language}. Empirically, we find in Figure \ref{fig:flores-bbc-comparison} that such a bias is either very close to zero.

\begin{figure*}[t]
    \centering
    \vspace{-1cm}
    \vspace{-1cm}
    \begin{subfigure}[b]{0.48\textwidth}
        \centering
        \includegraphics[width=\textwidth]{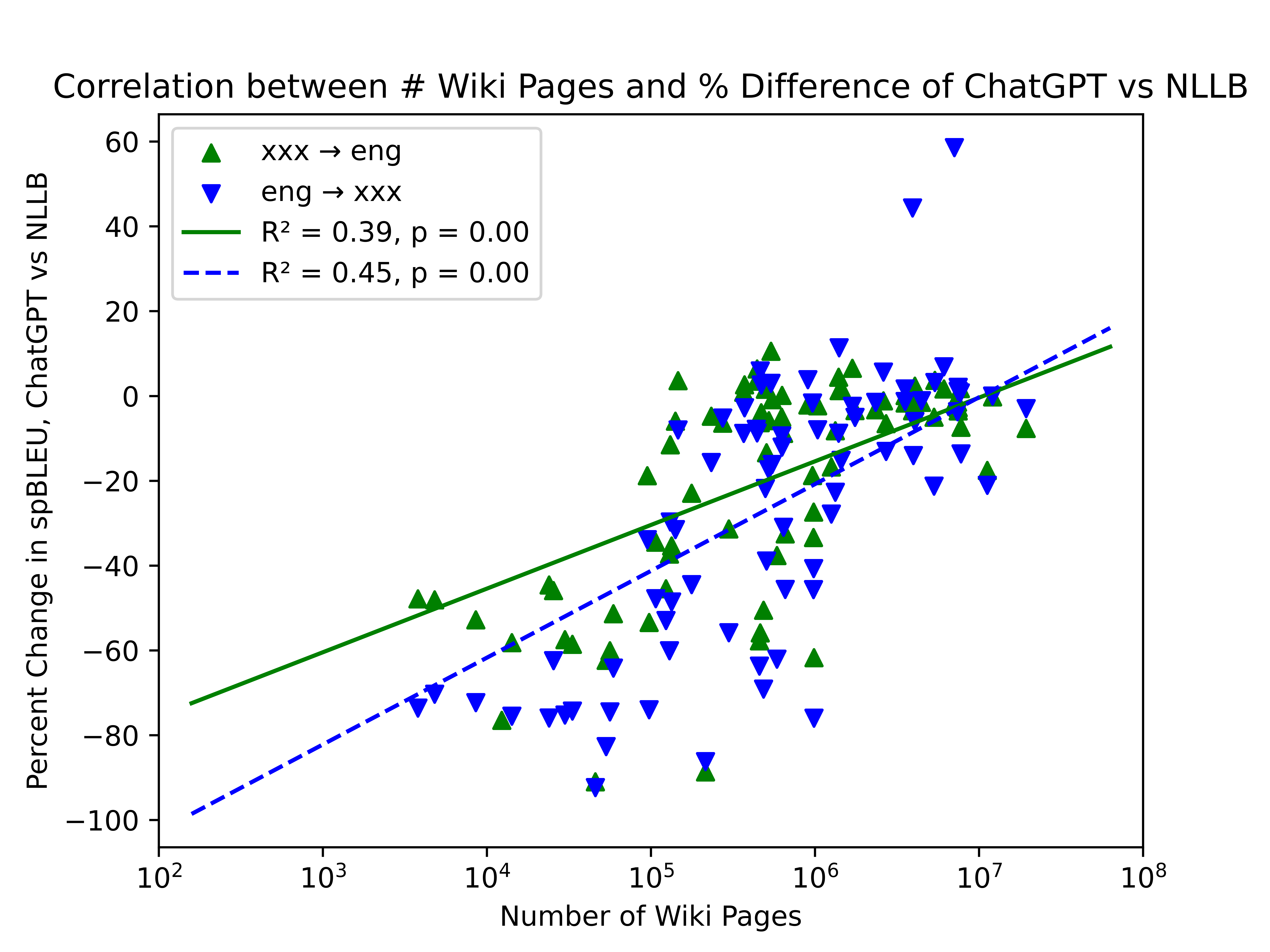}
        \caption{Correlations with ChatGPT translation performance.}
    \end{subfigure}
    \hfill
    \begin{subfigure}[b]{0.48\textwidth}
        \centering
        \includegraphics[width=\textwidth]{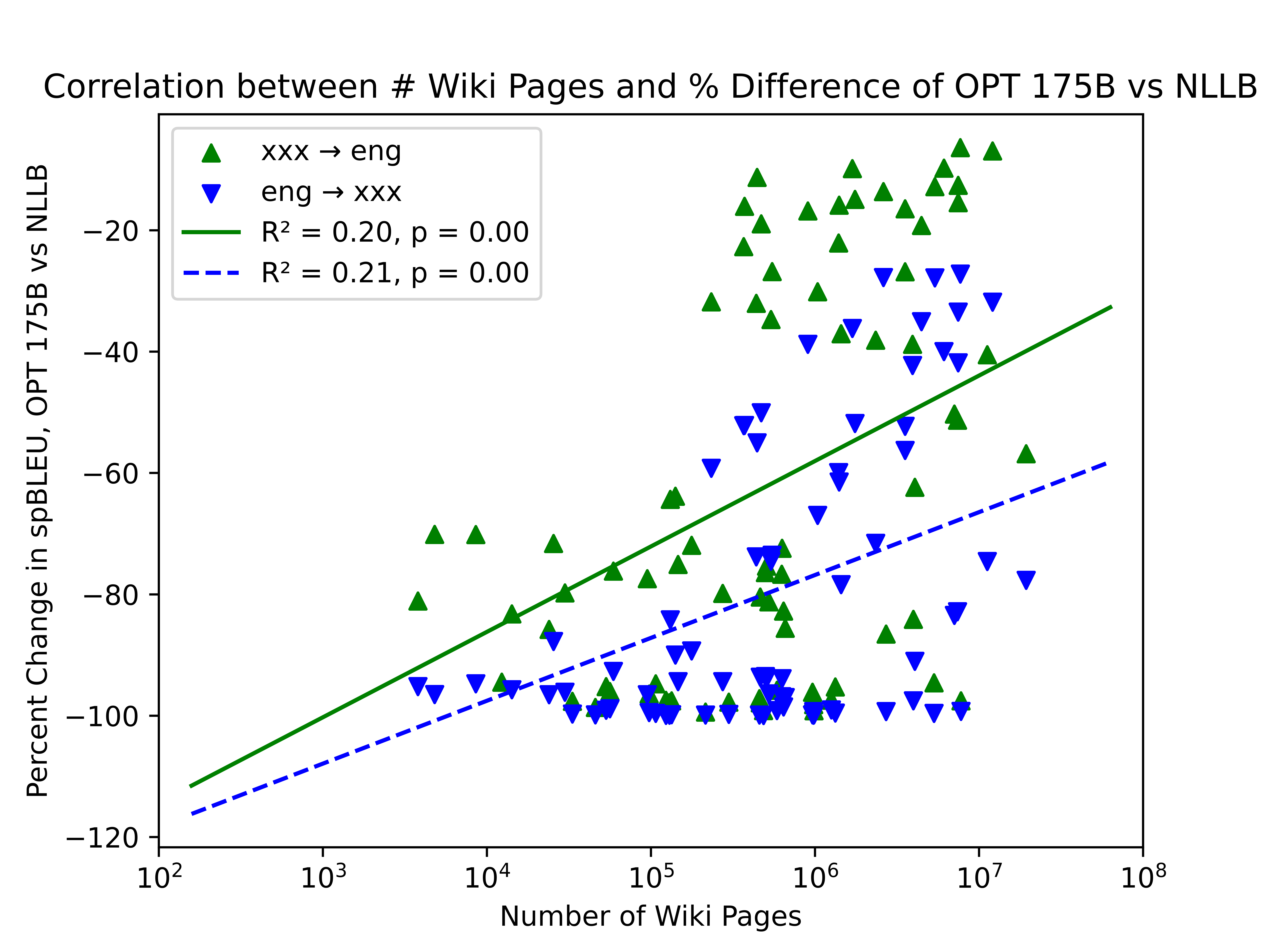}
        \caption{Correlations with OPT 175B translation performance.}
    \end{subfigure}
    
    \begin{subfigure}[b]{0.48\textwidth}
        \centering
        \includegraphics[width=\textwidth]{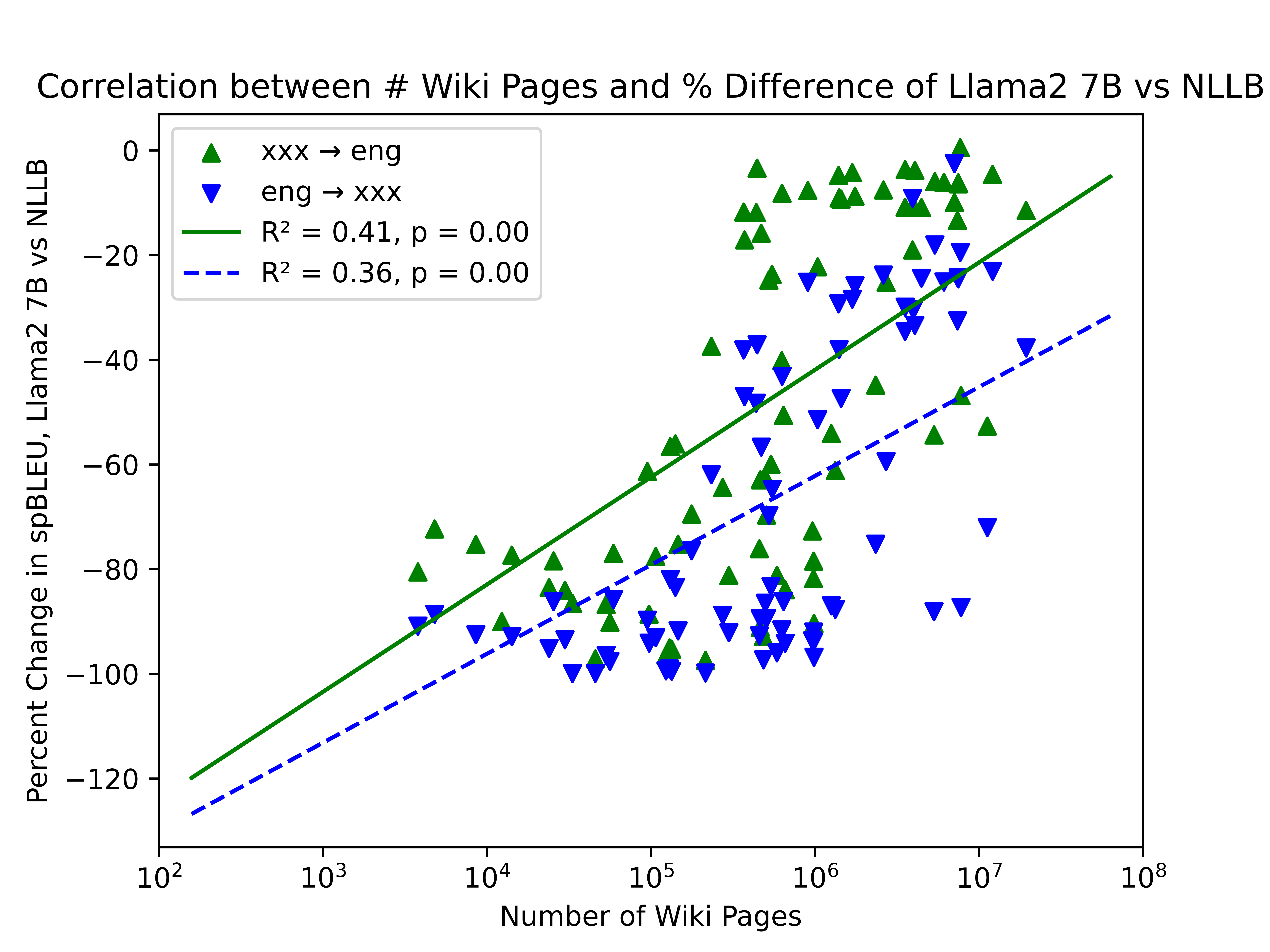}
        \caption{Correlations with LLAMA2 7B translation performance.}
    \end{subfigure}
    \hfill
    \begin{subfigure}[b]{0.48\textwidth}
        \centering
        \includegraphics[width=\textwidth]{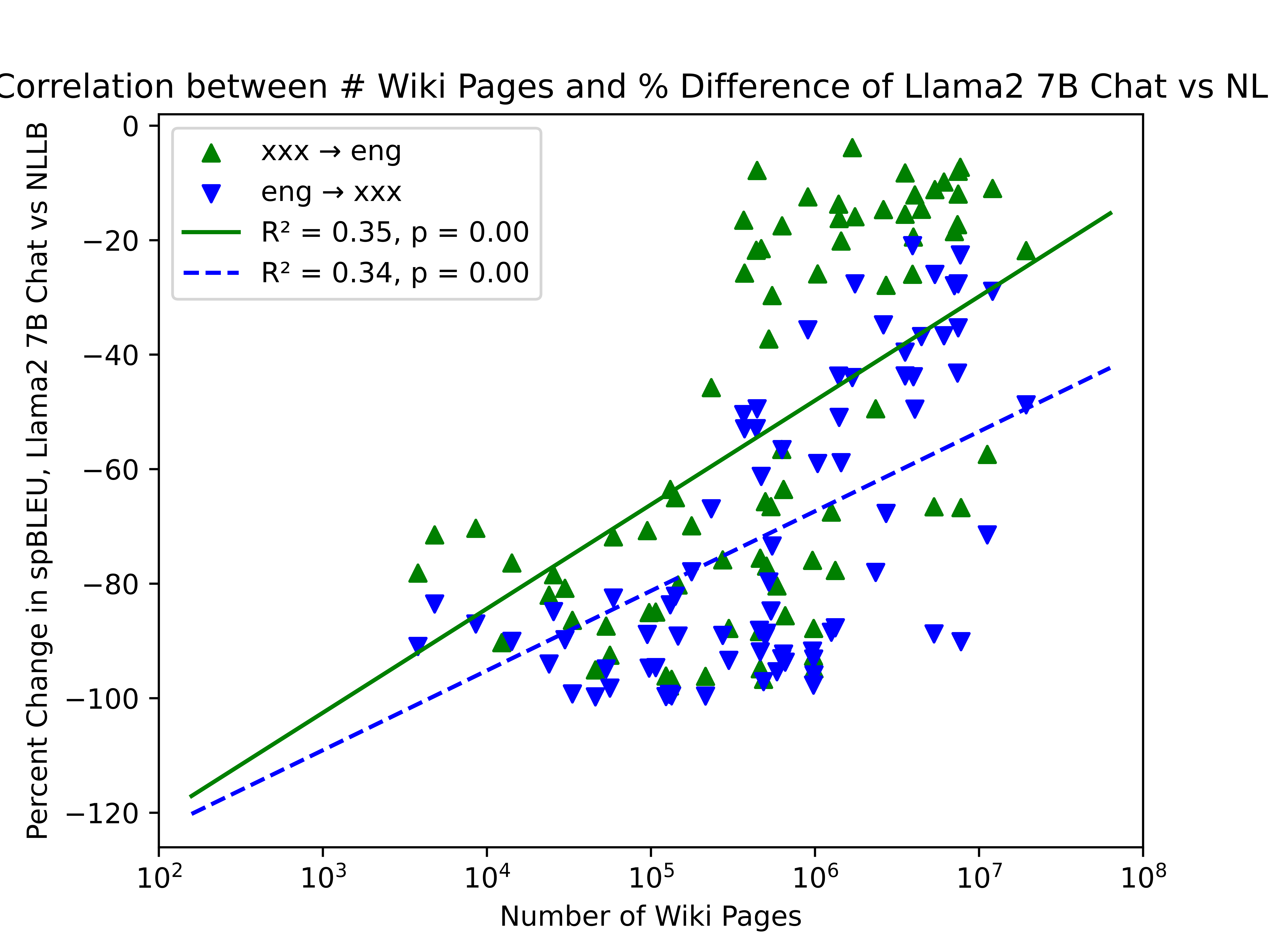}
        \caption{Correlations with LLAMA2 7B Chat translation performance.}
    \end{subfigure}
    \begin{subfigure}[b]{0.48\textwidth}
        \centering
        \includegraphics[width=\textwidth]{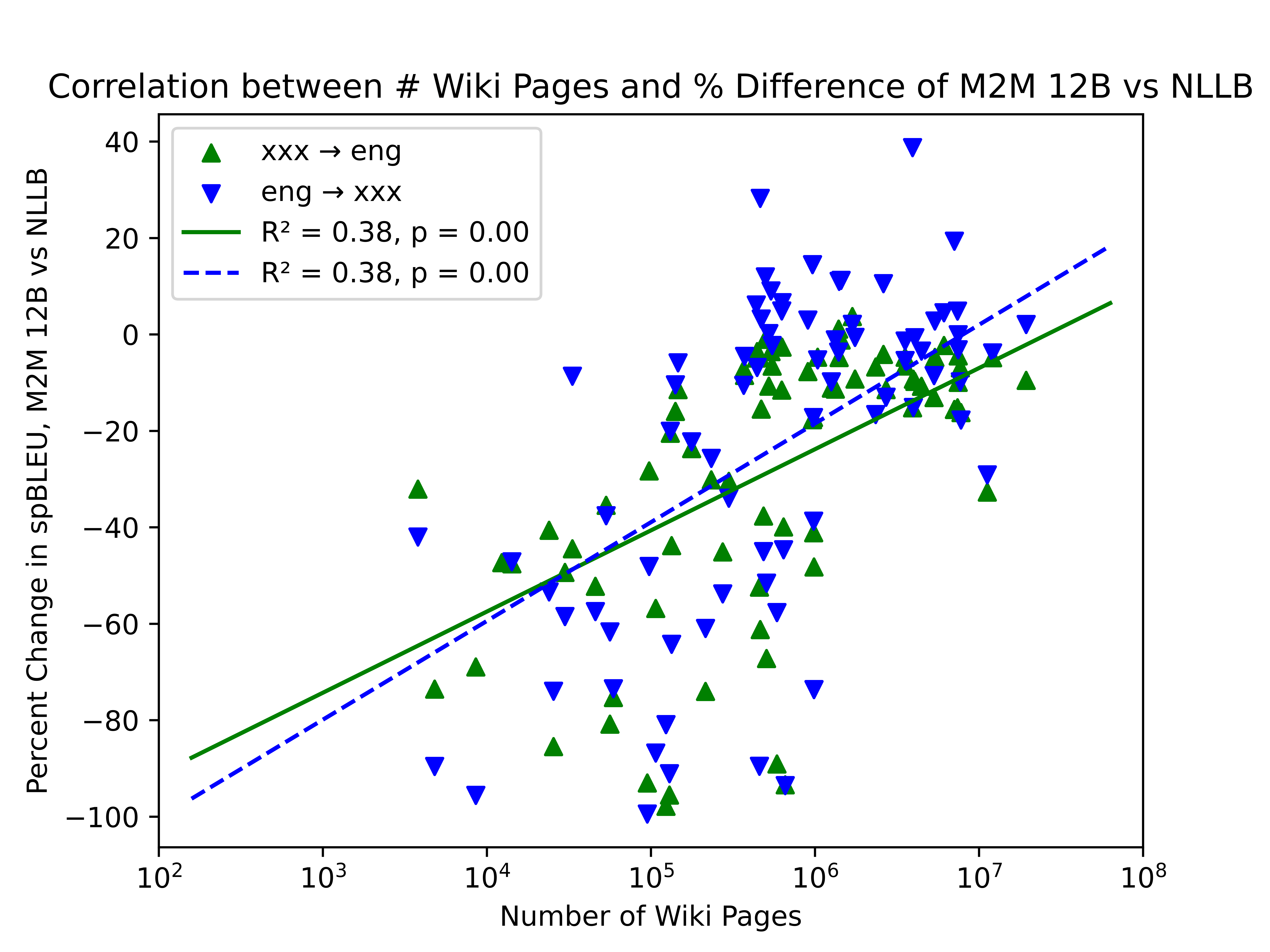}
        \caption{Correlations with M2M 12B translation performance.}
    \end{subfigure}
    \hfill
    \begin{subfigure}[b]{0.48\textwidth}
        \centering
        \includegraphics[width=\textwidth]{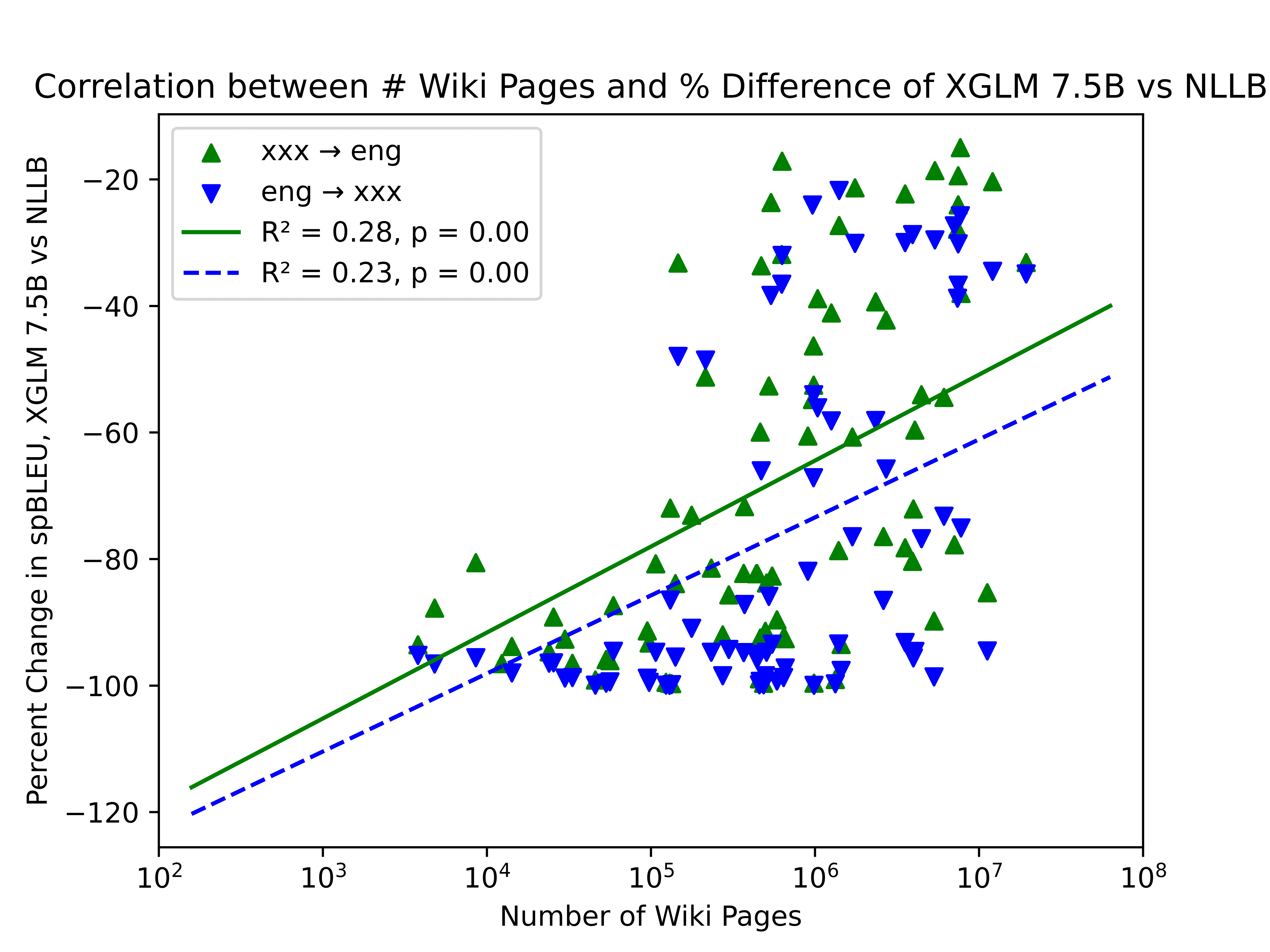}
        \caption{Correlations with XGLM 7.5B LLM translation performance.}
    \end{subfigure}
    \begin{subfigure}[b]{0.48\textwidth}
        \centering
        \includegraphics[width=\textwidth]{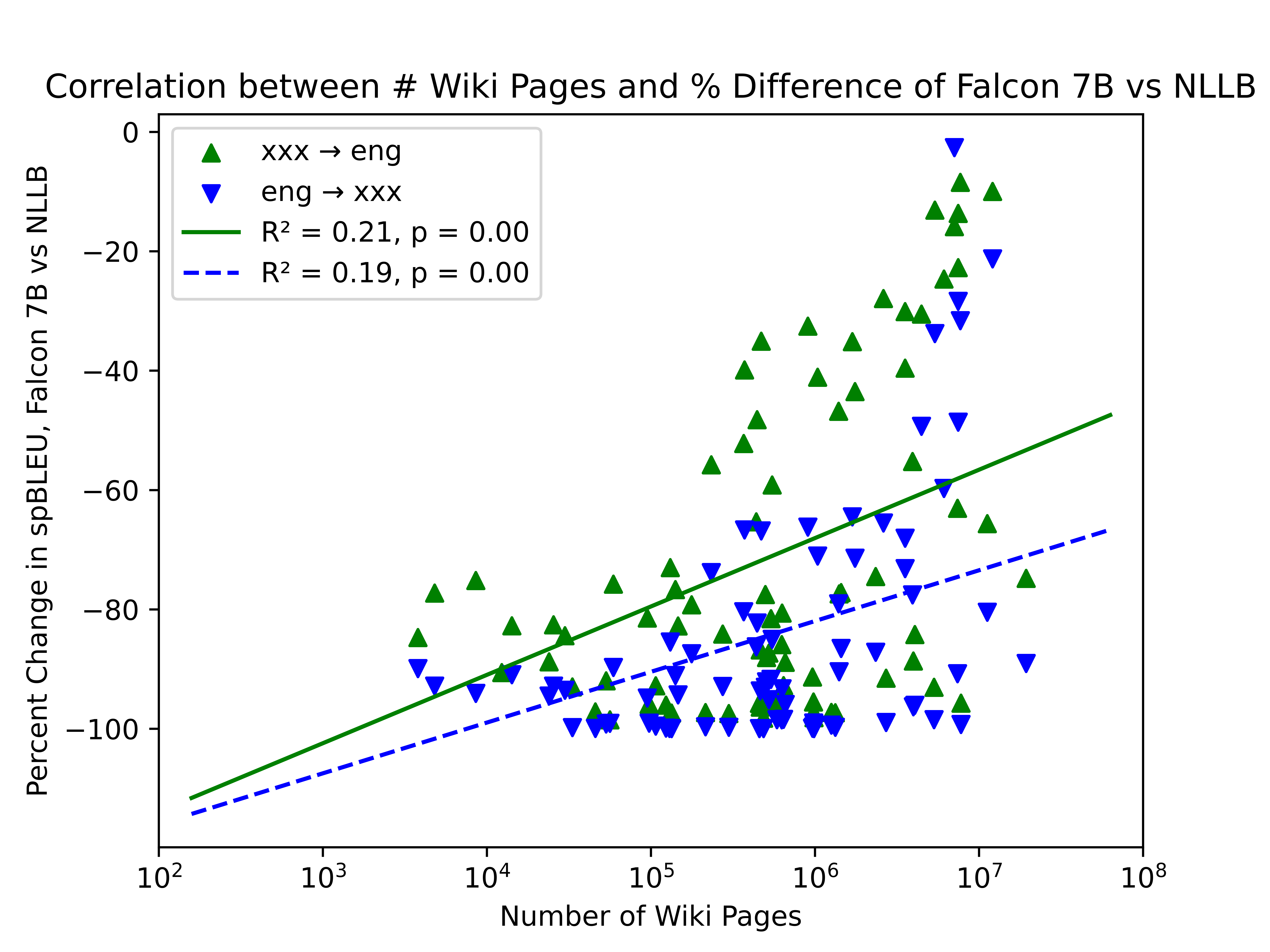}
        \caption{Correlations with Falcon 7B LLM translation performance.}
    \end{subfigure}
    \caption{Correlation plots for seven different LLMs, with data adapted from \citet{zhu2023multilingual}.}
    \label{fig:correlation-plots}
\end{figure*}

\begin{table*}[htbp]
\centering
\begin{tabular}{ll|ccc|ccc}
\hline
\multirow{2}{*}{Language} & \multirow{2}{*}{Code} & \multicolumn{3}{c|}{\texttt{xxx -> eng} spBLEU} & \multicolumn{3}{c}{\texttt{eng->xxx} spBLEU} \\
\cline{3-8}
 & & Google & Claude & NLLB & Google & Claude & NLLB \\
\hline
Amharic & amh\_Ethi & \textbf{43.18} & 42.95 & 41.74 & \textbf{32.98} & 26.05 & 29.30\\
Modern Standard Arabic & arb\_Arab & 48.21 & 49.04 & \textbf{49.91} & \textbf{48.18} & 44.40 & 42.60\\
Azerbaijani & azj\_Latn & 31.38 & \textbf{33.24} & 30.71 & 25.18 & \textbf{27.49} & 25.24\\
Bengali & ben\_Beng & 41.31 & \textbf{42.11} & 40.99 & \textbf{37.28} & 33.92 & 33.91\\
French & fra\_Latn & 49.96 & \textbf{50.82} & 48.94 & \textbf{55.52} & 54.46 & 51.17\\
Western Central Oromo & gaz\_Latn & \textbf{32.34} & 27.20 & 29.56 & \textbf{14.90} & 10.03 & 12.46\\
Gujarati & guj\_Gujr & \textbf{48.05} & 45.44 & 47.06 & \textbf{40.25} & 34.23 & 37.42\\
Hausa & hau\_Latn & \textbf{41.77} & 38.18 & 39.97 & \textbf{29.96} & 20.74 & 29.60\\
Hindi & hin\_Deva & \textbf{47.72} & 46.81 & 45.76 & \textbf{45.20} & 38.67 & 41.98\\
Igbo & ibo\_Latn & \textbf{39.03} & 32.90 & 37.18 & \textbf{23.26} & 17.71 & 20.55\\
Indonesian & ind\_Latn & 52.42 & \textbf{53.55} & 48.95 & \textbf{52.52} & 48.47 & 48.69\\
Japanese & jpn\_Jpan & \textbf{35.92} & 35.00 & 35.40 & 33.33 & \textbf{34.08} & 18.63\\
Kinyarwanda & kin\_Latn & 41.41 & \textbf{41.48} & 40.59 & \textbf{34.87} & 21.47 & 27.03\\
Kyrgyz & kir\_Cyrl & \textbf{32.99} & 32.07 & 27.79 & \textbf{29.57} & 27.76 & 27.97\\
Korean & kor\_Hang & \textbf{38.07} & 37.26 & 35.51 & \textbf{30.93} & 28.88 & 24.15\\
Marathi & mar\_Deva & \textbf{45.34} & 44.87 & 44.75 & \textbf{30.45} & 28.48 & 27.85\\
Burmese & mya\_Mymr & 34.27 & 25.57 & \textbf{34.67} & \textbf{24.66} & 23.62 & 18.26\\
Nepali & npi\_Deva & 48.72 & \textbf{50.53} & 48.59 & \textbf{38.27} & 34.49 & 29.83\\
Southern Pashto & pbt\_Arab & 41.68 & \textbf{43.30} & 40.62 & 23.71 & 21.11 & \textbf{25.59}\\
Persian & pes\_Arab & 44.46 & \textbf{48.35} & 43.93 & \textbf{39.39} & 38.53 & 35.42\\
Portuguese & por\_Latn & 59.55 & \textbf{59.89} & 58.29 & \textbf{59.96} & 59.48 & 54.99\\
Russian & rus\_Cyrl & 43.74 & \textbf{43.89} & 43.23 & \textbf{47.05} & 44.16 & 42.59\\
Sinhala & sin\_Sinh & 43.49 & \textbf{43.88} & 40.43 & \textbf{39.65} & 35.35 & 35.37\\
Somali & som\_Latn & 37.94 & \textbf{37.97} & 33.95 & \textbf{18.70} & 17.31 & 18.65\\
Spanish & spa\_Latn & 38.37 & \textbf{41.59} & 41.08 & \textbf{35.71} & 34.92 & 33.41\\
Swahili & swh\_Latn & 53.98 & \textbf{55.47} & 50.78 & \textbf{43.88} & 40.03 & 36.47\\
Tamil & tam\_Taml & 40.74 & 41.62 & \textbf{41.88} & \textbf{39.84} & 33.88 & 37.53\\
Telegu & tel\_Telu & \textbf{47.66} & 45.82 & 47.54 & \textbf{46.05} & 36.99 & 41.78\\
Thai & tha\_Thai & 34.16 & \textbf{42.13} & 39.80 & \textbf{46.11} & 44.52 & 33.33\\
Tigrinya & tir\_Ethi & 27.48 & 27.34 & \textbf{28.93} & 17.34 & 13.99 & \textbf{18.71}\\
Turkish & tur\_Latn & 47.79 & \textbf{49.40} & 45.52 & \textbf{45.21} & 42.12 & 41.30\\
Ukrainian & ukr\_Cyrl & 47.40 & \textbf{47.52} & 45.32 & \textbf{42.80} & 42.74 & 37.01\\
Urdu & urd\_Arab & 43.42 & \textbf{44.97} & 44.42 & \textbf{33.24} & 30.67 & 31.37\\
Northern Uzbek & uzn\_Latn & 45.25 & \textbf{46.43} & 39.51 & \textbf{36.33} & 34.50 & 31.09\\
Vietnamese & vie\_Latn & \textbf{42.17} & 41.12 & 40.46 & \textbf{45.29} & 41.65 & 42.48\\
Yoruba & yor\_Latn & 25.49 & \textbf{30.21} & 26.62 & 13.09 & \textbf{15.37} & 12.80\\
\hline
\end{tabular}
\caption{spBLEU scores on the FLORES-200 evaluation set of the Google Translate model versus the Claude 3 Opus model versus the NLLB-54B model. \textbf{Warning}: There is evidence for data contamination of Claude on this evaluation set in Section \ref{sec:results}.}
\label{tab:flores-translation-bleu}
\end{table*}

\begin{table*}[htbp]
\centering
\begin{tabular}{ll|ccc|ccc}
\hline
\multirow{2}{*}{Language} & \multirow{2}{*}{Code} & \multicolumn{3}{c|}{\texttt{xxx->eng} chrF++} & \multicolumn{3}{c}{\texttt{eng->xxx} chrF++} \\
\cline{3-8}
 & & Google & Claude & NLLB & Google & Claude & NLLB \\
\hline
Amharic & amh\_Ethi & \textbf{62.64} & 61.79 & 60.61 & \textbf{41.42} & 35.58 & 37.89\\
Modern Standard Arabic & arb\_Arab & 67.38 & \textbf{68.34} & \textbf{68.34} & \textbf{61.23} & 58.20 & 56.88\\
Azerbaijani & azj\_Latn & 55.55 & \textbf{57.76} & 54.82 & 44.45 & \textbf{46.20} & 43.33\\
Bengali & ben\_Beng & 62.29 & \textbf{62.98} & 61.09 & \textbf{50.43} & 48.08 & 46.99\\
French & fra\_Latn & 68.90 & \textbf{70.28} & 67.26 & \textbf{69.47} & 68.97 & 66.25\\
Western Central Oromo & gaz\_Latn & \textbf{52.73} & 49.46 & 50.41 & \textbf{39.86} & 34.91 & 37.28\\
Gujarati & guj\_Gujr & \textbf{67.66} & 64.53 & 66.79 & \textbf{55.77} & 51.01 & 53.18\\
Hausa & hau\_Latn & \textbf{59.55} & 56.97 & 57.64 & \textbf{52.56} & 45.67 & 52.10\\
Hindi & hin\_Deva & \textbf{66.50} & 65.00 & 65.51 & \textbf{60.24} & 55.35 & 57.92\\
Igbo & ibo\_Latn & \textbf{57.36} & 53.62 & 56.04 & \textbf{44.80} & 39.78 & 42.06\\
Indonesian & ind\_Latn & 70.12 & \textbf{70.39} & 67.66 & \textbf{71.35} & 68.79 & 68.77\\
Japanese & jpn\_Jpan & \textbf{58.32} & 58.04 & 56.26 & \textbf{35.89} & 33.07 & 26.77\\
Kinyarwanda & kin\_Latn & 59.74 & \textbf{60.07} & 58.46 & \textbf{56.82} & 45.87 & 49.79\\
Kyrgyz & kir\_Cyrl & 55.08 & \textbf{55.31} & 50.10 & \textbf{47.69} & 46.11 & 45.93\\
Korean & kor\_Hang & \textbf{59.46} & 59.20 & 56.65 & \textbf{38.70} & 36.29 & 33.89\\
Marathi & mar\_Deva & \textbf{65.48} & 64.04 & 64.77 & \textbf{49.15} & 47.55 & 45.87\\
Burmese & mya\_Mymr & \textbf{56.26} & 48.87 & 55.79 & 39.93 & \textbf{40.66} & 31.45\\
Nepali & npi\_Deva & 67.80 & \textbf{68.12} & 66.98 & \textbf{55.77} & 52.80 & 46.05\\
Southern Pashto & pbt\_Arab & 62.08 & \textbf{62.93} & 61.15 & 40.33 & 37.70 & \textbf{41.42}\\
Persian & pes\_Arab & 64.94 & \textbf{66.94} & 63.62 & 53.96 & \textbf{55.48} & 49.61\\
Portuguese & por\_Latn & 75.14 & \textbf{75.39} & 74.11 & \textbf{73.71} & 73.49 & 70.33\\
Russian & rus\_Cyrl & 64.13 & \textbf{65.41} & 63.13 & \textbf{62.08} & 60.61 & 57.86\\
Sinhala & sin\_Sinh & \textbf{63.62} & 63.56 & 60.54 & \textbf{49.92} & 47.73 & 42.65\\
Somali & som\_Latn & 57.35 & \textbf{57.43} & 53.61 & \textbf{43.62} & 42.00 & 43.09\\
Spanish & spa\_Latn & 60.49 & \textbf{63.56} & 61.56 & \textbf{56.73} & 56.40 & 54.50\\
Swahili & swh\_Latn & 70.34 & \textbf{70.77} & 67.80 & \textbf{63.87} & 61.14 & 58.24\\
Tamil & tam\_Taml & \textbf{61.59} & 61.49 & 61.14 & \textbf{56.04} & 52.62 & 54.39\\
Telegu & tel\_Telu & \textbf{66.02} & 64.40 & 65.89 & \textbf{59.52} & 52.26 & 56.06\\
Thai & tha\_Thai & 58.25 & \textbf{62.87} & 60.36 & \textbf{50.80} & 50.26 & 43.32\\
Tigrinya & tir\_Ethi & \textbf{51.68} & 50.60 & 50.69 & 25.70 & 23.00 & \textbf{26.05}\\
Turkish & tur\_Latn & 66.62 & \textbf{68.06} & 64.40 & \textbf{60.91} & 58.87 & 57.76\\
Ukrainian & ukr\_Cyrl & 65.72 & \textbf{66.45} & 63.97 & 59.21 & \textbf{59.86} & 54.95\\
Urdu & urd\_Arab & 64.02 & \textbf{65.79} & 63.96 & \textbf{50.58} & 49.11 & 49.56\\
Northern Uzbek & uzn\_Latn & 64.64 & \textbf{66.63} & 60.02 & \textbf{55.26} & 54.31 & 51.46\\
Vietnamese & vie\_Latn & 61.65 & \textbf{61.82} & 60.54 & \textbf{60.77} & 59.21 & 59.00\\
Yoruba & yor\_Latn & 45.99 & \textbf{49.23} & 46.57 & \textbf{38.27} & 30.19 & 38.25\\
\hline
\end{tabular}
\caption{chrF++ scores on the FLORES-200 evaluation set of the Google Translate model versus the Claude 3 Opus model versus the NLLB-54B model. \textbf{Warning}: There is evidence for data contamination of Claude on this evaluation set as argued in Section \ref{sec:results}.}
\label{tab:flores-translation-chrf}
\end{table*}

\begin{table*}[htbp]
\centering
\begin{tabular}{ll|ccc|ccc}
\hline
\multirow{2}{*}{Language} & \multirow{2}{*}{Code} & \multicolumn{3}{c|}{\texttt{xxx->eng} spBLEU} & \multicolumn{3}{c}{\texttt{eng->xxx} spBLEU} \\
\cline{3-8}
 & & Google & Claude & NLLB & Google & Claude & NLLB \\
\hline
Amharic & amh\_Ethi & 17.13 & \textbf{17.55} & 17.33 & \textbf{16.49} & 12.04 & 16.05\\
Modern Standard Arabic & arb\_Arab & 45.49 & 44.59 & \textbf{46.49} & \textbf{71.04} & 49.48 & 47.12\\
Azerbaijani & azj\_Latn & \textbf{35.42} & 34.41 & 34.66 & \textbf{48.31} & 33.80 & 38.03\\
Bengali & ben\_Beng & 25.28 & \textbf{26.92} & 25.37 & \textbf{35.39} & 24.64 & 32.06\\
French & fra\_Latn & 56.51 & \textbf{57.30} & 54.91 & \textbf{72.90} & 61.41 & 66.71\\
Western Central Oromo & gaz\_Latn & 20.02 & 17.07 & \textbf{20.55} & 11.63 & 7.48 & \textbf{11.67}\\
Gujarati & guj\_Gujr & \textbf{31.06} & 28.48 & 30.37 & \textbf{32.59} & 22.69 & 25.36\\
Hausa & hau\_Latn & \textbf{27.43} & 25.73 & 26.76 & \textbf{28.51} & 16.76 & 23.35\\
Hindi & hin\_Deva & 27.64 & \textbf{28.59} & 26.89 & \textbf{22.65} & 20.88 & 21.74\\
Igbo & ibo\_Latn & \textbf{41.95} & 33.82 & 38.20 & \textbf{58.13} & 27.16 & 34.90\\
Indonesian & ind\_Latn & \textbf{50.63} & 48.37 & 48.76 & \textbf{71.74} & 47.76 & 54.90\\
Japanese & jpn\_Jpan & \textbf{24.96} & 21.66 & 21.72 & \textbf{18.75} & 14.82 & 11.00\\
Kinyarwanda & kin\_Latn & 28.24 & 28.70 & \textbf{30.39} & 14.76 & 12.18 & \textbf{17.26}\\
Kyrgyz & kir\_Cyrl & 21.49 & \textbf{21.56} & 19.44 & \textbf{23.11} & 18.69 & 21.69\\
Korean & kor\_Hang & 20.49 & 20.62 & \textbf{22.67} & 19.82 & \textbf{20.29} & 17.70\\
Marathi & mar\_Deva & \textbf{18.89} & 16.94 & 18.26 & \textbf{24.14} & 15.17 & 19.53\\
Burmese & mya\_Mymr & \textbf{27.41} & 25.37 & 23.55 & \textbf{26.28} & 14.91 & 19.86\\
Nepali & npi\_Deva & 25.93 & \textbf{27.24} & 26.44 & \textbf{25.35} & 18.23 & 23.69\\
Southern Pashto & pbt\_Arab & 28.98 & 30.68 & \textbf{34.29} & \textbf{37.20} & 24.06 & 29.24\\
Persian & pes\_Arab & \textbf{38.12} & 36.55 & 37.56 & \textbf{43.36} & 35.96 & 32.04\\
Portuguese & por\_Latn & \textbf{54.85} & 53.59 & 53.55 & \textbf{67.49} & 58.45 & 56.99\\
Russian & rus\_Cyrl & 34.10 & 37.14 & \textbf{37.34} & \textbf{44.03} & 33.65 & 34.63\\
Sinhala & sin\_Sinh & \textbf{38.62} & 35.06 & 36.17 & \textbf{39.62} & 28.85 & 33.17\\
Somali & som\_Latn & \textbf{49.72} & 46.12 & 44.30 & \textbf{74.11} & 32.50 & 40.53\\
Spanish & spa\_Latn & \textbf{52.77} & 46.36 & 49.89 & \textbf{63.85} & 53.09 & 54.56\\
Swahili & swh\_Latn & \textbf{56.87} & 50.71 & 47.63 & \textbf{83.18} & 50.85 & 50.60\\
Tamil & tam\_Taml & \textbf{29.56} & 25.32 & 27.61 & \textbf{47.29} & 28.41 & 35.80\\
Telegu & tel\_Telu & 19.83 & \textbf{21.06} & 18.56 & \textbf{23.14} & 16.24 & 19.56\\
Thai & tha\_Thai & 19.21 & 19.18 & \textbf{20.07} & 26.58 & \textbf{27.31} & 23.64\\
Tigrinya & tir\_Ethi & \textbf{36.45} & 31.28 & 31.53 & \textbf{58.09} & 17.01 & 27.84\\
Turkish & tur\_Latn & 28.93 & 30.64 & \textbf{31.41} & \textbf{26.94} & 24.74 & 26.10\\
Ukrainian & ukr\_Cyrl & 38.45 & \textbf{39.67} & 37.60 & \textbf{53.47} & 44.83 & 42.71\\
Urdu & urd\_Arab & 38.18 & 36.64 & \textbf{39.78} & \textbf{48.87} & 38.14 & 40.15\\
Northern Uzbek & uzn\_Latn & 31.55 & \textbf{32.75} & 26.72 & \textbf{0.12} & 0.09 & 0.04\\
Vietnamese & vie\_Latn & \textbf{40.92} & 39.44 & 38.06 & \textbf{54.43} & 40.24 & 42.77\\
Yoruba & yor\_Latn & 20.59 & 21.15 & \textbf{21.53} & 13.75 & 15.49 & \textbf{18.27}\\
\hline
\end{tabular}
\caption{spBLEU scores on the BBC evaluation set of the Google Translate model versus the Claude 3 Opus model versus the NLLB-54B model.}
\label{tab:bbc-translation-bleu}
\end{table*}

\begin{table*}[htbp]
\centering
\begin{tabular}{ll|ccc|ccc}
\hline
\multirow{2}{*}{Language} & \multirow{2}{*}{Code} & \multicolumn{3}{c|}{\texttt{xxx->eng} chrF++} & \multicolumn{3}{c}{\texttt{eng->xxx} chrF++} \\
\cline{3-8}
 & & Google & Claude & NLLB & Google & Claude & NLLB \\
\hline
Amharic & amh\_Ethi & \textbf{38.84} & 38.08 & 37.88 & \textbf{24.96} & 21.32 & 24.26\\
Modern Standard Arabic & arb\_Arab & \textbf{66.44} & 65.23 & 66.31 & \textbf{77.77} & 60.86 & 59.55\\
Azerbaijani & azj\_Latn & \textbf{56.72} & 56.56 & 55.77 & \textbf{63.03} & 53.44 & 55.50\\
Bengali & ben\_Beng & 45.55 & \textbf{46.66} & 46.14 & \textbf{50.14} & 42.77 & 47.66\\
French & fra\_Latn & 71.51 & \textbf{71.90} & 70.49 & \textbf{80.63} & 74.87 & 76.93\\
Western Central Oromo & gaz\_Latn & \textbf{41.35} & 40.25 & 40.36 & \textbf{34.83} & 30.12 & 34.08\\
Gujarati & guj\_Gujr & 50.28 & 48.43 & \textbf{50.44} & \textbf{49.27} & 41.89 & 41.61\\
Hausa & hau\_Latn & \textbf{47.62} & 45.88 & 47.38 & \textbf{49.30} & 40.07 & 45.78\\
Hindi & hin\_Deva & \textbf{48.88} & 48.62 & 48.44 & \textbf{44.69} & 43.63 & 43.56\\
Igbo & ibo\_Latn & \textbf{58.07} & 52.31 & 55.37 & \textbf{69.16} & 45.10 & 53.06\\
Indonesian & ind\_Latn & \textbf{66.65} & 64.95 & 65.25 & \textbf{85.70} & 72.25 & 74.40\\
Japanese & jpn\_Jpan & \textbf{46.42} & 44.76 & 44.09 & \textbf{26.41} & 23.21 & 17.66\\
Kinyarwanda & kin\_Latn & 48.11 & 49.30 & \textbf{49.89} & 38.14 & 35.79 & \textbf{40.29}\\
Kyrgyz & kir\_Cyrl & 41.90 & \textbf{42.49} & 39.70 & \textbf{43.26} & 40.61 & 42.38\\
Korean & kor\_Hang & 47.77 & \textbf{47.83} & 47.11 & 26.65 & \textbf{27.13} & 25.15\\
Marathi & mar\_Deva & \textbf{38.13} & 37.26 & 37.56 & \textbf{45.51} & 38.17 & 42.23\\
Burmese & mya\_Mymr & \textbf{49.91} & 48.47 & 46.28 & \textbf{37.77} & 31.13 & 34.23\\
Nepali & npi\_Deva & 49.13 & \textbf{49.27} & 48.61 & \textbf{42.80} & 38.54 & 40.35\\
Southern Pashto & pbt\_Arab & 51.37 & 53.16 & \textbf{54.63} & \textbf{49.67} & 40.20 & 43.71\\
Persian & pes\_Arab & 58.22 & 58.17 & \textbf{58.38} & \textbf{56.31} & 50.66 & 47.82\\
Portuguese & por\_Latn & \textbf{71.25} & 70.84 & 70.66 & \textbf{79.05} & 72.73 & 71.86\\
Russian & rus\_Cyrl & 55.69 & \textbf{57.45} & 55.83 & \textbf{56.11} & 50.04 & 51.32\\
Sinhala & sin\_Sinh & \textbf{57.64} & 54.50 & 55.11 & \textbf{48.54} & 39.45 & 40.75\\
Somali & som\_Latn & \textbf{66.15} & 63.27 & 62.48 & \textbf{82.03} & 54.20 & 60.50\\
Spanish & spa\_Latn & \textbf{69.02} & 66.22 & 66.89 & \textbf{75.94} & 69.66 & 69.60\\
Swahili & swh\_Latn & \textbf{70.28} & 66.25 & 64.18 & \textbf{89.28} & 69.40 & 68.71\\
Tamil & tam\_Taml & \textbf{49.62} & 47.15 & 48.26 & \textbf{63.75} & 50.69 & 55.02\\
Telegu & tel\_Telu & 40.09 & \textbf{41.35} & 39.30 & \textbf{38.41} & 33.79 & 35.78\\
Thai & tha\_Thai & \textbf{40.52} & 40.39 & 40.16 & 33.16 & \textbf{37.67} & 33.24\\
Tigrinya & tir\_Ethi & \textbf{55.28} & 50.52 & 51.03 & \textbf{63.25} & 25.98 & 34.69\\
Turkish & tur\_Latn & 50.80 & 51.41 & \textbf{51.84} & \textbf{49.22} & 47.91 & 48.19\\
Ukrainian & ukr\_Cyrl & 56.21 & \textbf{57.70} & 55.55 & \textbf{64.55} & 58.56 & 56.04\\
Urdu & urd\_Arab & 59.28 & 59.80 & \textbf{59.89} & \textbf{61.58} & 53.71 & 55.11\\
Northern Uzbek & uzn\_Latn & \textbf{51.36} & 51.14 & 46.23 & \textbf{2.27} & 2.21 & 2.20\\
Vietnamese & vie\_Latn & \textbf{59.66} & 59.05 & 57.26 & \textbf{66.94} & 58.06 & 59.13\\
Yoruba & yor\_Latn & 41.77 & \textbf{43.78} & 42.53 & 26.03 & 27.95 & \textbf{28.47}\\
\hline
\end{tabular}
\caption{chrF++ scores on the BBC evaluation set of the Google Translate model versus the Claude 3 Opus model versus the NLLB-54B model.}
\label{tab:bbc-translation-chrf}
\end{table*}

\end{document}